\documentclass[runningheads]{llncs}
\usepackage{eccv}
\usepackage{eccvabbrv}
\usepackage{graphicx}
\usepackage{booktabs}
\usepackage{epsfig}
\usepackage{amsmath}
\usepackage{amssymb}
\usepackage{lipsum}
\usepackage{multirow}
\usepackage{adjustbox}
\usepackage{color}
\usepackage{colortbl}
\usepackage[accsupp]{axessibility}

\usepackage{orcidlink}

\def\Ie{\emph{I.e}\onedot}
\def\cf{\emph{c.f}\onedot} \def\Cf{\emph{C.f}\onedot}
\def\etc{\emph{etc}\onedot} \def\vs{\emph{vs}\onedot}
\def\wrt{w.r.t\onedot} 
\def\aka{a.k.a\onedot} 
\def\dof{d.o.f\onedot}
\def\etal{\emph{et al}\onedot}
\def\ie{\textit{i.e.}}
\def\eg{\textit{e.g.}}

\definecolor{gg}{RGB}{84, 130, 53}
\definecolor{pp}{RGB}{153, 51, 255}
\definecolor{bb}{RGB}{53, 121, 184}

\begin{document}

\title{DreamView: Injecting View-specific Text Guidance into Text-to-3D Generation} 

\titlerunning{DreamView}

\author{Junkai Yan\inst{1,2,3,\dag}\orcidlink{0009-0009-6531-0070} \and
Yipeng Gao\inst{1,2,3,\dag} \and
Qize Yang\inst{3} \and
Xihan Wei\inst{3} \and
Xuansong Xie\inst{3} \and
Ancong Wu\inst{1,2,}\thanks{: Corresponding authors are A.Wu and WS.Zheng. \dag~: Equal contribution.\\This work was done when J. Yan and Y. Gao were interns at Alibaba.}\orcidlink{0000-0002-7969-3190} \and
Wei-Shi Zheng\inst{1,2,}$^\star$\orcidlink{0000-0001-8327-0003}}

\authorrunning{J.Yan et al.}

\institute{$^1$ School of Computer Science and Engineering, Sun Yat-sen University, China;\\
$^2$ Key Laboratory of Machine Intelligence and Advanced Computing, Ministry of Education, China; $^3$Institute for Intelligent Computing, Alibaba Group\\
\email{\{yanjk3,gaoyp23\}@mail2.sysu.edu.cn, \{qize.yqz,xihan.wxh,xingtong.xxs\}@alibaba-inc.com, wuanc@mail.sysu.edu.cn, wszheng@ieee.org}}

\maketitle

\begin{figure}
    \centering
    \includegraphics[width=1\linewidth]{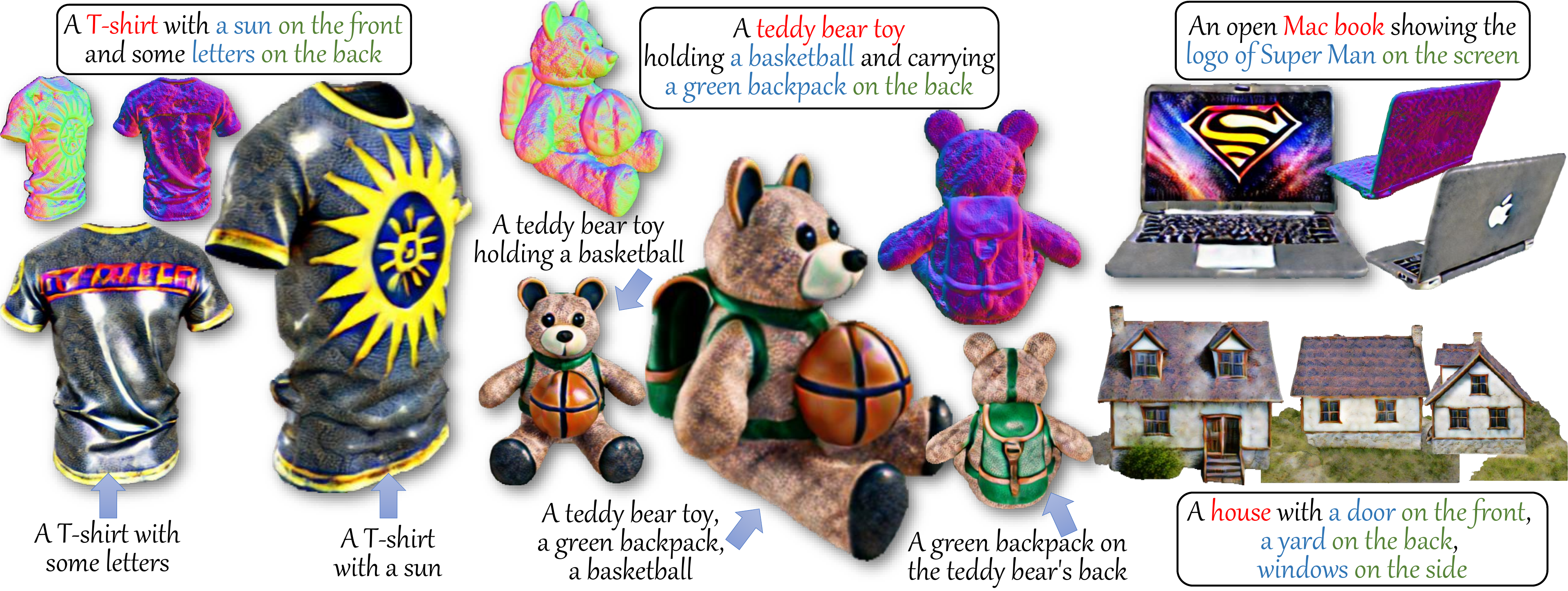}
    \caption{\textbf{Text-to-3D generation} (RGB images and normal maps) of our \textbf{DreamView}, where users can control what and where to generate via providing an overall text description (surrounded by a black box on the figure) and view-specific texts, thus achieving customizable 3D generation. The \textcolor{red}{subject}, \textcolor{bb}{object}, and the \textcolor{gg}{position} of the object in the overall text are marked in \textcolor{red}{red}, \textcolor{bb}{blue}, and \textcolor{gg}{green}, respectively. For the two generated results on the right, we only show the overall text.}
    \vspace{-2.5em}
    \label{fig-intro}
\end{figure}

\begin{abstract}
  Text-to-3D generation, which synthesizes 3D assets according to an overall text description, has significantly progressed. However, a challenge arises when the specific appearances need customizing at designated viewpoints but referring solely to the overall description for generating 3D objects. For instance, ambiguity easily occurs when producing a T-shirt with distinct patterns on its front and back using a single overall text guidance. In this work, we propose \textbf{DreamView}, a text-to-image approach enabling multi-view customization while maintaining overall consistency by adaptively injecting the view-specific and overall text guidance through a collaborative text guidance injection module, which can also be lifted to 3D generation via score distillation sampling. DreamView is trained with large-scale rendered multi-view images and their corresponding view-specific texts to learn to balance the separate content manipulation in each view and the global consistency of the overall object, resulting in a dual achievement of customization and consistency. Consequently, DreamView empowers artists to design 3D objects creatively, fostering the creation of more innovative and diverse 3D assets. Code and model will be released at \href{https://github.com/iSEE-Laboratory/DreamView}{here}.
  \keywords{Generative model \and Text-to-image generation \and Text-to-3D generation}
\end{abstract}

\section{Introduction}
The surge in demand for diverse 3D asset creation spans many domains, including robotics simulation~\cite{robotics_simulation_1,robotics_simulation_2}, vison recognition with synthesis~\cite{video_action_1,video_action_2,video_gen_1,video_gen_2,video_gen_3,recognition_sys_1,recognition_sys_2,recognition_sys_3}, and architecture design~\cite{building_design_1,pointe,shape}, especially with the advancement of virtual and augmented reality technologies~\cite{survey1,survey2,survey3}. 
Despite the broadening scope of application, massively producing professional-grade 3D content necessitates artistic sensibility coupled with specialized skills in 3D modeling. 
Recent 3D synthesis works~\cite{0123,dreamfusion,sjc,fantasia3d,pointe,magic123,dmtet,neus,shape} attempt to produce high-quality 3D assets without much labor effort. Among them, text-to-3D generation~\cite{dreamfusion,pointe,pdreamer,mvdream,sweetdream,shape} has garnered considerable attention for its ability to create 3D assets from text prompts, utilizing text-2D prior for text-3D representation learning~\cite{ulip,ulip-2,openshape,mixcon3d,uni3d,tamm}. 

\begin{figure}[t]
    \centering
    \includegraphics[width=1\linewidth]{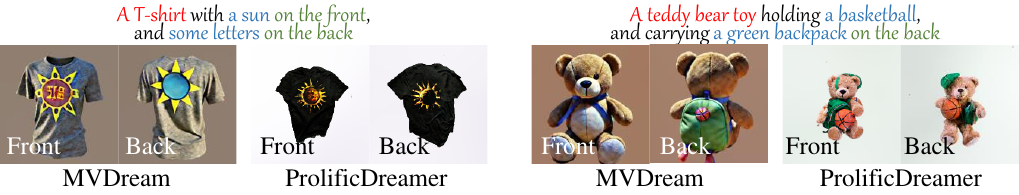}
    \caption{Text-to-3D generation (front and back views) of recent works, which cannot generate content strictly aligned with texts while may suffer from inconsistent problems.}
    \label{fig-motivation}
\end{figure}

Existing text-to-3D works can be divided into two streams: one conducts a direct generation~\cite{pointe,shape}, and another adopts 2D pre-trained text-to-image models to optimize differentiable 3D representations~\cite{dreamfusion,sjc,fantasia3d}, known as the 2D-lifting method. The latter has shown a promising ability to produce {high-fidelity} 3D assets. However, in the evolving landscape of text-to-3D generation, a notable limitation of existing methods is their reliance on a shared text description among all views of a generated 3D object. Therefore, the inherent diversity in views that a single 3D object can present is generally overlooked, such as varying patterns on different views of a T-shirt, as shown in~\Cref{fig-intro}. Consequently, their approaches are inadequate for customizing viewpoints in 3D instances, which restricts their usability to meet tailored or complicated requirements, as shown in~\Cref{fig-motivation}.

In this work, we propose \textbf{DreamView} for customizable text-to-3D generation. We achieve this by constructing a highly customizable and consistent text-to-image model that can synthesize specific image views of an object according to the provided overall and view-specific texts, where the overall text describes the object from the global level and the view-specific text only contains contents appearing in a specific viewpoint, as shown by the samples in~\Cref{fig-intro}. In DreamView, the overall text is shared among views to promote \textbf{consistent} image generation across viewpoints. Moreover, the view-specific text serves its corresponding view for \textbf{customizable} image view generation. To balance these two kinds of text guidance rather than letting one of them dominate, we propose an adaptive text guidance injection module, which acts on each block of the diffusion model and dynamically selects which text should be used in the current block, achieving a dynamic balance between consistency and customization.

DreamView is trained on a large-scale rendered dataset containing multi-view images with paired texts to seek the ability to generate customizable 2D multi-view images from varying viewpoints and maintain instance-level consistency across views. More importantly, this ability can easily transfer to 3D generation via 2D-lifting methods~\cite{dreamfusion,sjc}, facilitating a consistent and customizable text-to-3D synthesis. The dual achievement of viewpoint customization and 3D instance-level consistency marks a significant advancement in the realm of text-to-3D generation, making a modest step in this field.

We present several impressive results in~\Cref{fig-intro}, where our DreamView generates high-quality 3D assets that faithfully adhere to the text prompts, showcasing unique appearances defined by each view-specific text. Additionally, we conduct qualitative and quantitative comparisons with other methods regarding text-to-image generation and text-to-3D instance generation. Moreover, we conduct a user study to analyze user preferences. The results collectively highlight the abilities of DreamView to produce customizable and consistent 3D objects.

\section{Related Works}
\noindent\textbf{Text-to-image generation.}
With the proposal of several large-scale text-image datasets ~\cite{laion5b,laion400m,Conceptualcaptions,wit}, diffusion models~\cite{sdm1,sdm2,ddim,ddpm} have become increasingly popular in 2D image generation. Diffusion models consist of a forward process that gradually adds noise to data and a reverse (sampling) process that denoises pure noise. By leveraging various conditions such as text and mask, diffusion models can generate high fidelity and diverse content faithful to the user-provided prompts in image editing~\cite{imagic,insp2p}, inpainting~\cite{smartbrush,repaint}, \etc. 
Recently, diffusion models have shown their potential capabilities in 3D generation~\cite{dreamfusion,sjc} due to the strong 2D {generation performance}. Compared to previous models, ours absorbs the strong viewpoint consistency and customization from a 3D rendered dataset via an adaptive text guidance injection module, making it more suitable for 3D synthesis.

\vspace{0.7em}
\noindent\textbf{2D-lifting text-to-3D generation.}
Recently, how to exploit powerful text-to-image generative models~\cite{sdm1,sdm2,ddim,ddpm} to perform text-to-3D synthesis has received considerable attention \cite{sjc,pdreamer,dreamfusion,magic3d,mvdream}. A notable contribution in this field is the SDS (score distillation sampling) proposed in DreamFusion~\cite{dreamfusion}, utilizing diffusion priors as score functions to supervise the optimization of 3D representation. Additionally, Wang \etal~\cite{sjc} apply the chain rule on the learned gradients of a diffusion model and back-propagate the score of it through the Jacobian of a differentiable render. Subsequently, a series of efforts were made to improve generation quality~\cite{fantasia3d,dreamtime,pdreamer} and ensure 3D consistency~\cite{debias,perpneg,mvdream,sweetdream,3dfuse,mvdiff}. 
Despite these methods achieving impressive 3D generation, they still struggle to follow the provided texts strictly, failing to customize the appearance of 3D objects while may also encounter inconsistent problems, as shown in~\Cref{fig-motivation}. To overcome these challenges, we propose to utilize view-specific text guidance via an adaptive text guidance injection module to introduce customization ability while adaptively maintaining the instance-level consistency, thus empowering a more creative text-to-3D generation.

\section{DreamView}
\label{sec-dreamview}
Current 2D-lifting text-to-3D generation models predominantly consider a view-shared text prompt, termed `overall text', to guide generation. However, this simplified pipeline has a significant limitation: encapsulating all appearance attributes of an object within a single text prompt complicates the customization of specific viewpoints. These models facilitate only object-level guidance over the object's generation, lacking the capability to delineate the object's appearance from various viewpoints. For example, as shown in~\Cref{fig-motivation}, although the text prompts have provided explicit requirements on two views, the current 3D generative models can neither generate letters on the T-shirt nor place the basketball and backpack in the correct position.

In this work, we explore customizing the appearance of the generated 3D object from different views, thereby generating more imaginative 3D assets. We achieve this by making use of the view-specific text, which describes the appearance of an object from a specific viewpoint. Because the view-specific text is not shared among views, viewpoint inconsistency may occur sometimes (see~\Cref{sec-exp-2d}). To overcome this challenge, we propose an adaptive guidance injection module to adaptively collaborate the overall and the view-specific text to achieve a dynamic balance between consistent and customizable generation. The proposed injection module is plugged into a text-to-image model and can be subsequently lifted into 3D generation via score distillation sampling~\cite{dreamfusion}. To distinguish the text-to-image and text-to-3D variants of DreamView, we denote them DreamView-2D (\Cref{fig-dreamview-2d}) and DreamView-3D (\Cref{fig-dreamview-3d}), respectively.

\subsection{DreamView-2D for Text-to-image Generation}
\label{sec-dreamview-2d}

\begin{figure}[t]
    \centering
    \includegraphics[width=1\linewidth]{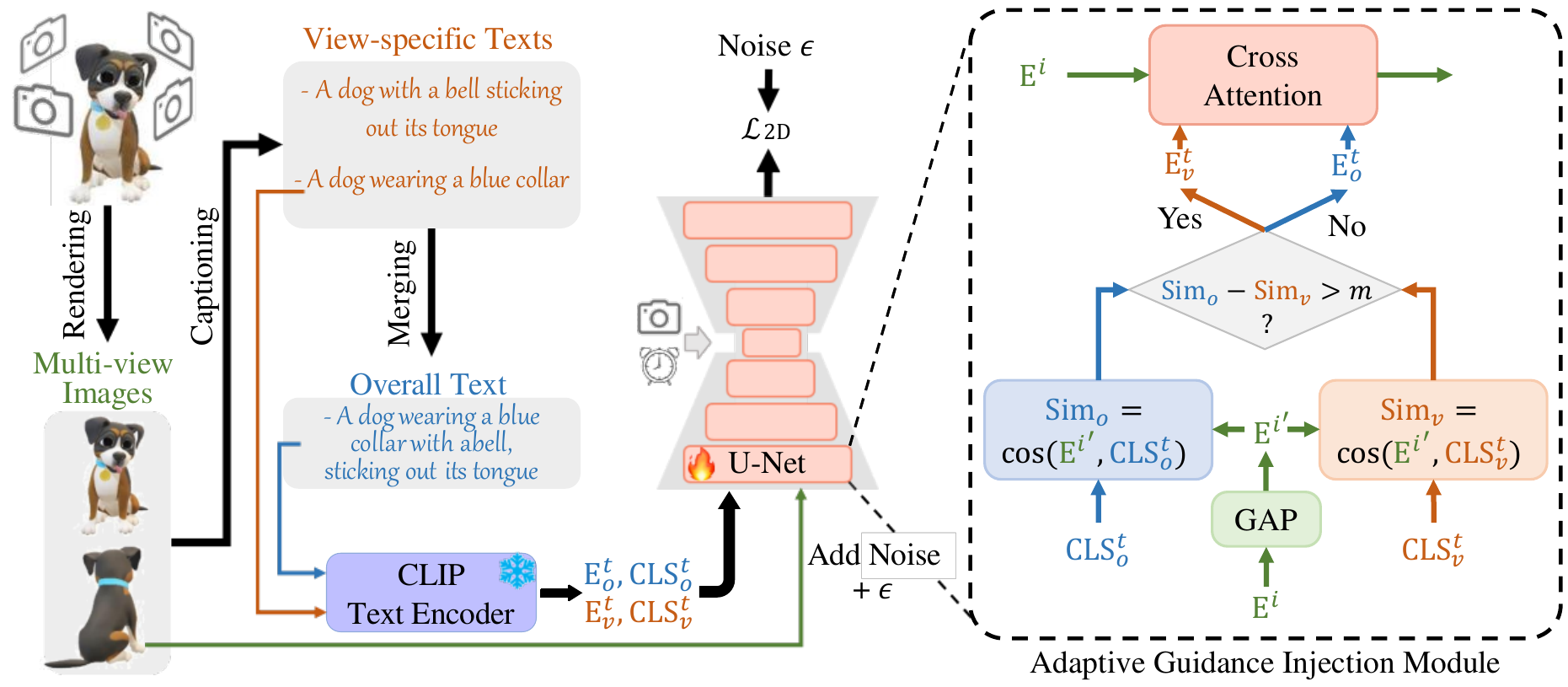}
    \caption{The overall framework of DreamView-2D. \textbf{Left:} the data preparation pipeline and the data flow of DreamView-2D. The raw 3D objects from the Objaverse dataset~\cite{objv} are first rendered to multi-view images and captioned by BLIP-2~\cite{blip2}. Finally, the view-specific texts are merged by GPT-4~\cite{gpt4} to form the overall text. With this image-text paired data, DreamView-2D, augmented by an adaptive guidance inject module, is trained to learn a trade-off between 3D consistency and customization. \textbf{Right:} the detail of the adaptive guidance injection module working in each U-Net block of the model, which measures the similarity between the image embedding and the two types of text embedding to determine which text guidance should be used in the current U-Net block, thus achieving an adaptive balance between the consistency and customization. A margin hyper-parameter is used in the model to control the balance.}
    \label{fig-dreamview-2d}
\end{figure}

\noindent\textbf{Data preparation.}
To facilitate view-specific customization, we construct a 3D dataset for training, including paired multi-view images with view-specific texts and the overall text, as shown in the left side of~\Cref{fig-dreamview-2d}. Our dataset is sourced from the Objaverse dataset~\cite{objv}, which only contains a raw text description for each object and lacks view-specific texts. Thus, it is inadequate to achieve our intention of injecting view-specific texts. To complement the required view-specific text, we utilize advanced large multi-modal models to generate high-quality view-specific and overall text descriptions for each 3D asset in the Objaverse dataset. 

Specifically, the construction process is divided into three steps: rendering, captioning, and merging. (1) In the rendering step, each 3D asset is first densely rendered as several multi-view images with $512\times512$ resolution, and the corresponding camera poses are saved. (2) The captioning stage is carried out via the BLIP-2~\cite{blip2} model, which is applied to generate a caption for each rendered image, forming the `view-specific text' for training our DreamView-2D. (3) In the final step, merging, all of the view-specific texts from different views of a 3D object are consolidated by GPT-4~\cite{gpt4}, forming the `overall text description' of this 3D asset. More dataset details are in the \textit{supplement material}.

\vspace{0.7em}
\noindent\textbf{Adaptive guidance injection module.} 
As mentioned before, we propose an adaptive guidance injection module to balance the guidance from the overall and view-specific texts, whose core idea is determining which guidance dominates the current diffusion U-Net block~\cite{sdm1,unet} and injecting the other guidance into the cross-attention layer as the condition to achieve a dynamic balance, as shown in the right side of~\Cref{fig-dreamview-2d}.

Specifically, we denote the text embeddings of the overall text prompt and view-specific text prompts given by the CLIP text encoder~\cite{clip} as $\mathrm{E}^{t}_o$ and $\mathrm{E}^{t}_v$, respectively, where the superscript $^t$ denotes `text embedding'. Similarly, their global representations, \ie, class tokens are also output by the text encoder of CLIP, denoted as $\mathrm{CLS}^{t}_o$ and $\mathrm{CLS}^{t}_v$. The image embedding is denoted as $\mathrm{E}^{i}$. In each U-Net block, we measure the similarity between the image and text embeddings via 
\begin{equation}
\text{Sim}=\cos(\text{GAP}(\mathrm{E}^{i}),\mathrm{CLS}^{t}),
\label{eq-sim}
\end{equation}
where the `GAP' denotes the global average pooling operation converting the image feature map to a representation vector, and `cos' indicates the cosine similarity operator. Applying this equation to $\mathrm{CLS}^{t}_o$ and $\mathrm{CLS}^{t}_v$, we obtain the image-text similarity of the overall text (Sim$_o$) and the view-specific text (Sim$_v$). The magnitude between these two similarity values determines which guidance will be injected into the current block. For example, a larger Sim$_o$ suggests that the current $\mathrm{E}^{i}$ absorbs more overall guidance than view-specific guidance, and thus we inject the view-specific one. This process can be formulated as:
\begin{equation}
\text{Guidance}=
\begin{cases}
    \mathrm{E}^{t}_v, & \text{if Sim}_o - \text{Sim}_v > m  \\
    \mathrm{E}^{t}_o, & \text{else}
\end{cases},
\label{eq-guidance}
\end{equation}
where $m$ is a margin hyper-parameter to control the propensity to consistency or customization. Intuitively, a large margin means more overall guidance will be used, resulting in stronger 3D consistency. Conversely, customization will be dominant with a small margin.

Through the above modeling, we convert the trade-off problem between consistency and customization to adjust the margin, thus achieving an adaptive balance between these two properties in a text-to-image model and subsequently benefiting the text-to-3D generation.

\vspace{0.7em}
\noindent\textbf{Optimization.}
Given dataset $\mathcal{D}=\{(x_r, T_r^O)\}_1^N$, where the $N$ is the number of rendered 3D assets in the dataset, the $T^O$ is the overall text description of the 3D asset $x$. Besides, each 3D asset $x$ is formulated as $\{I_s, T_s, c_s\}_1^M$, representing the rendered image, its text caption given by BLIP-2, and the corresponding rendered camera pose, respectively, and the $M$ denotes the number of rendered views. With these image-text-camera pairs, we can exploit overall and view-specific texts to jointly guide the generative model via the proposed adaptive guidance injection module for seeking consistency and customization capabilities.

In the training phase, the input of the diffusion U-Net can be denoted as $(\mathbf{x}; y, c, t)$, where $\mathbf{x}$, $c$ and $t$ are the image, the camera position, and the sampled time step, respectively. Besides, $y=\{\mathrm{E}^{t}_o, \mathrm{E}^{t}_v, \mathrm{CLS}^{t}_o, \mathrm{CLS}^{t}_v\}$, and it is prepared for the injection module mentioned in the previous paragraph to provide the condition for each U-Net block according to~\Cref{eq-sim,eq-guidance}. Overall, the U-Net parameterized by $\theta$ is trained by optimizing the diffusion loss:
\begin{equation}
\mathcal{L}_{2D}(\theta,\mathcal{D}) = \mathbb{E}_{\mathbf{x},y,c,t,\epsilon}[\|\epsilon-\epsilon_\theta(\mathbf{x}_t;y,c,t)\|^2_2],
\end{equation}
where the $\epsilon$ and $\epsilon_\theta$ are the ground truth and predicted random noise, and $\mathbf{x}_t$ represents the noisy image. In practice, DreamView-2D is trained to generate single or multiple image views for 3D objects with the above loss function. To preserve the identity of the 3D object from multiple viewpoints, an expanded attention mechanism that models the relation among views~\cite{mvdiff,mvdream,wu2023tune,blattmann2023align,liu2023syncdreamer} will be applied when generating multiple views for a 3D object. Through the above optimization, DreamView-2D is expected to learn a trade-off between consistency and customization with the help of the adaptive guidance injection module.

\vspace{0.7em}
\noindent\textbf{Inference.}
Unlike the training phase, where each image is paired with a view-specific text and an overall text, users must provide both texts during inference, resulting in a heavy burden of writing text prompts, especially with the increase of generated views. For example, if users hope to generate $n$ image views of a 3D object, they must provide $n$ view-specific texts and one overall text, which is unacceptable. Fortunately, considering the spatial continuity of 3D objects, whose appearance usually does not change intensively within a continuous range of viewpoints, we can roughly divide the object into four views: front, back, left, and right. In this case, only five texts are needed, \ie, one overall text and four view-specific texts, reducing the burdens of users. In addition, one can draw support from large language models to write view-specific texts, and thus, only an overall text is required, which will be discussed in the \textit{supplementary materials}. Furthermore, if users do not require customization, our model can also adopt an overall text to conduct generation.

\subsection{DreamView-3D for Text-to-3D Generation}
\label{sec-dreamview-3d}
\begin{figure}[t]
    \centering
    \includegraphics[width=0.85\linewidth]{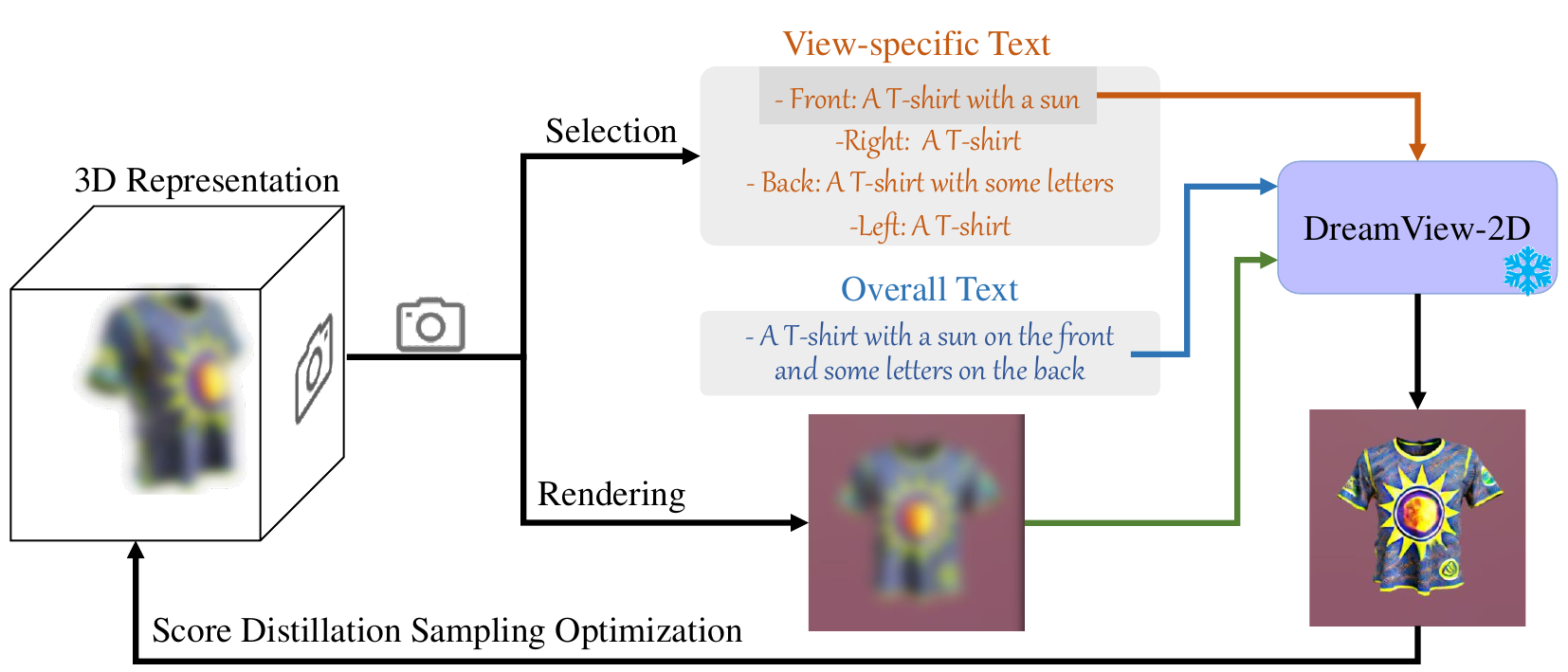}
    \caption{The overall framework of DreamView-3D, which optimizes a 3D representation via score distillation sampling~\cite{dreamfusion} supervised by DreamView-2D, thus inheriting the consistent and customizable priors.}
    \label{fig-dreamview-3d}
\end{figure}

Following typical 2D-lifting text-to-3D generation methods~\cite{sjc,dreamfusion,pdreamer,fantasia3d}, we adopt our DreamView-2D model as a teacher model and distill the priors to supervise 3D generation. We build our DreamView-3D based on the score distillation sampling~(SDS) technique proposed by DreamFusion~\cite{dreamfusion}, which is supposed to introduce the trade-off ability between consistency and customization in DreamView-2D into 3D generation. To this end, we replace the used text-to-image generation model in DreamFusion with our DreamView-2D. In addition, we also need to introduce view-specific texts together.

\vspace{0.7em}
\noindent\textbf{Applying view-specific text.}
Similar to the inference stage discussed in~\Cref{sec-dreamview-2d}, we divide the azimuth angle from 0-360 degrees into four parts, corresponding to the front, right, left, and back, respectively. Four view-specific texts are associated with four non-intersecting intervals spanning the azimuth angle from 0-360 degrees, respectively. Then, once the camera position for rendering the 3D representation is given, we can determine which view-specific text to use in the current viewpoint. Thus, our DreamView-2D with the adaptive guidance injection module can be easily plugged into 2D-lifting 3D generation techniques.

\vspace{0.7em}
\noindent\textbf{Overall pipeline.}
Our DreamView-3D is illustrated in~\Cref{fig-dreamview-3d}. Firstly, a camera position $c$ is sampled, and a 3D representation $\phi$ is projected to a 2D image $\mathbf{x}$ via differentiable rendering $g$ with the camera position. The camera azimuth is used to select view-specific text embedding, which is then fed into the injection module together with the overall text embedding and their class tokens. The diffusion model $\theta$ accepts the rendered image with noise $\mathbf{x}_t$, the text condition $y=\{\mathrm{E}^{t}_o, \mathrm{E}^{t}_v, \mathrm{CLS}^{t}_o, \mathrm{CLS}^{t}_v\}$, the camera position $c$, and the sampled time $t$ as inputs, outputting an estimated $\hat{\mathbf{x}}_0$, which is formulated as $\hat{\mathbf{x}}_0 = \epsilon_\theta(\mathbf{x}_t;y,c,t)$. The 3D representation is then optimized by an $\mathbf{x}_0$-reconstruction loss~\cite{mvdream}:
\begin{equation}
\label{eq-dreamview-2d}
\mathcal{L}_{3D}(\phi,\mathbf{x}=g(\phi)) = \mathbb{E}_{c,t,\epsilon}[\|\mathbf{x}-\hat{\mathbf{x}}_0\|^2_2],
\end{equation}
where the gradient of the diffusion model $\theta$ is detached to distill its priors into the differentiable 3D representation~\cite{dreamfusion}.

Through Equation~\ref{eq-dreamview-2d}, the 3D representation can inherit the powerful view content customization and instance-level consistency capabilities in our DreamView-2D, thus achieving customizable and consistent text-to-3D generation.

\section{Experiments}
\label{sec-exp}
\subsection{Implementation Details}
\label{sec-exp-imp}
\noindent\textbf{DreamView-2D.}
We train the DreamView-2D by combining our 3D rendered dataset and the 2D LAION dataset~\cite{laion5b} to ensure 3D consistency, customization, and visual quality. The model is initialized by the SD-v2.1 model~\cite{sdm1} and is trained on 16 V100 GPUs with a total batch size of 2,048. The learning rate is set to 1e$^{-4}$. For each 3D object, we randomly select four orthogonal image views and resize them to $256\times256$ to train the model, and the corresponding camera positions are normalized into a sphere. Regarding the inference, we adopt the DDIM~\cite{ddim} sampler with 50 sampling steps and a classifier-free guidance (CFG) scale of 7.5 for generating four image views simultaneously. The margin is randomly sampled from -0.1 to 0.1 during training and fixed at -0.025 during inference. Moreover, we also evaluate the model with our validation set, where we measure the CLIP-text score and CLIP-image score between the generated images and their corresponding texts and ground truth images, as well as the inception score for quality judgment.

\vspace{0.7em}
\noindent\textbf{DreamView-3D.}
We implement DreamView-3D by building upon threestudio~\cite{threestudio} and substituting Stable Diffusion~\cite{sdm1} in DreamFusion~\cite{dreamfusion} with our DreamView-2D for text-to-3D generation. To represent the 3D content, we employ the implicit-volume approach~\cite{barron2021mip} and optimize it for 10,000 steps using the AdamW optimizer~\cite{adamw} with a learning rate of 0.01. During optimization, SDS's maximum and minimum time steps are linearly annealed. Initially, the rendering resolution is set to 64$\times$64 for the first 5,000 steps and is then increased to 256$\times$256. After 5,000 steps, we enable soft shading~\cite{magic3d}. In most cases, we divide the azimuth angle of the camera position into four intervals: $[10, 170]$ is the front side, $(170, 190)$ is the right side, $[190, 350]$ is the back side, and the remaining part is the left side. The margin is set as -0.025. For further details, including the view-specific prompts, please refer to the \textit{supplement material}. 

\subsection{Text-to-Image Generation}
\label{sec-exp-2d}
In this section, we evaluate the image generation capabilities of DreamView-2D and conduct ablation studies on the hyper-parameter balancing the customization and consistency ability, \ie, the margin.

\begin{table}[t]
  \label{tab-comparison}
  \caption{\textbf{Comparisons on image synthesis quality.} The metrics are evaluated on the validation set of our rendered dataset. The numbers separated by the `/' in the line of SD-v2.1 and MVDream denote using overall or view text for generation, respectively. DreamView adaptively selects which text should be used.}
  \renewcommand{\arraystretch}{1.1}
  \centering
  \resizebox{0.75\linewidth}{!}{
    \begin{tabular}{l|c|c|c|c}
    \toprule
        \multirow{2}{*}{Method} & \multicolumn{3}{c|}{CLIP Score ($\uparrow$)} & ~Inception \\ 
        \cline{2-4}
        & ~Overall Text~ & ~View Text~ & ~GT Image~ & ~Score ($\uparrow$) \\ 
        \hline
        Ground Truth & 34.5 & 34.8 & 1.00 & 10.3 \\ 
        \hline
        SD-v2.1~\cite{sdm1} & 29.2/28.3 & 26.8/29.4 & 0.48/0.53 & 15.3/15.6 \\ 
        MVDream~\cite{mvdream} & 31.3/29.9 & 28.6/30.1 & 0.65/0.67 & 13.2/13.1 \\ 
        DreamView-2D~ & 31.1 & 32.1 & 0.73 & 14.5 \\
    \bottomrule
    \end{tabular}
  }
\end{table}

\vspace{0.7em}
\noindent\textbf{Quantitative comparison with other methods.} 
In~\Cref{tab-comparison}, we compare the image synthesis quality of DreamView-2D with other text-to-image generative models using a validation set of 1,000 objects from our dataset. We generate four image views for each object and show the average result of all generated images. We mainly focus on four metrics: the CLIP image-text score of the overall and view-specific text, the CLIP image-image score between the generated and the ground truth images, and the inception score. Intuitively, the CLIP score with overall text can roughly represent consistency as it is shared among views, and the CLIP score with view-specific text reflects the customization capability since it is specified in a specific view. The CLIP image-image score comprehensively reflects consistency and customization of the generated image, as the ground truth image is rendered from the 3D object. Besides, the inception score focuses on the visual quality of the image.

The results in~\Cref{tab-comparison} show that SD-v2.1~\cite{sdm1} achieves the best inception score and the worst CLIP score. It is only trained with 2D data and thus cannot generalize to 3D multi-view generation. Moreover, MVDream slightly outperforms ours in the overall text (+0.2) when it generates images according to the overall text. In contrast, its CLIP score with view text is significantly inferior to ours (-3.5), indicating a lack of customization capability. When the view text is applied to guide the image generation, the CLIP score with the view text of MVDream improved (+1.5) with certain consistency losses (-1.4) but still lower than ours. Regarding the CLIP image-image score, DreamView outperforms others clearly, indicating its good balance in consistency and customization. Lastly, the inception score of DreamView is slightly worse than SD-v2.1, but considering its consistency and customization, these losses are acceptable.

\begin{figure}[t]
  \centering
  \includegraphics[width=1\linewidth]{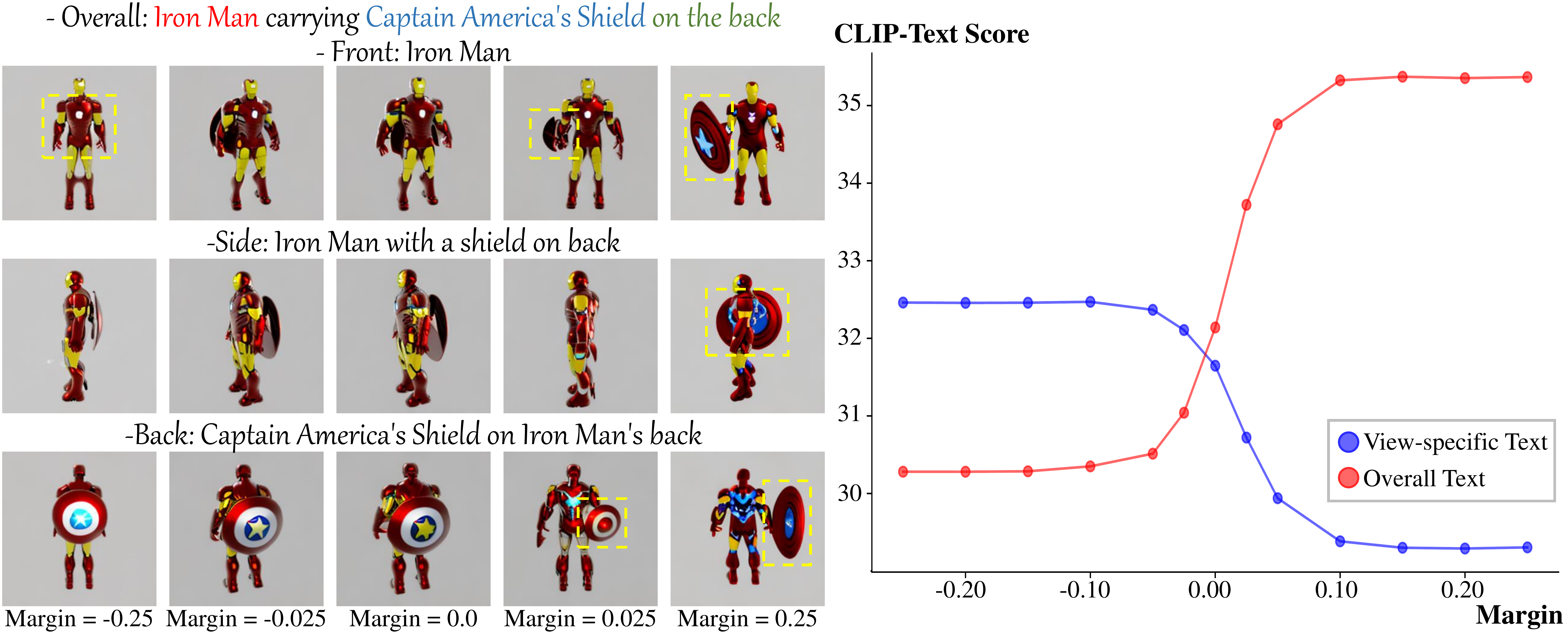}
  \caption{\textbf{Left:} qualitative text-to-image generation results of DreamView-2D with different margins. \textbf{Right:} quantitative evaluation results (CLIP image-text score) on the validation set with the margins change. As the margin gradually increases, customization will weaken while consistency will increase.}
  \label{fig-2d}
\end{figure}

\vspace{0.7em}
\noindent\textbf{Studies on the margin.}
The margin used in DreamView is designed to balance the guidance from the overall and view-specific text. We conduct qualitative and quantitative ablation studies in~\Cref{fig-2d} to validate this property. 

From the left figure, with the margin increases, one can observe a trend of being more consistent. Specifically, as shown by yellow boxes, the view-specific text for the front view does not specify `shield', resulting in the loss of the shield when the margin is small, which reflects stronger customization while weaker consistency. When the margin is large, \eg, 0.025 and 0.25, the shield no longer appears on Iron Man's back but on his arm, especially in the front view, demonstrating a better consistency but lacking customization. Besides, setting the margin to -0.025 and 0.0 shows a balanced trade-off between the two properties. By default, we set the margin to be -0.025 during inference.

Furthermore, the quantitative results on the right side of~\Cref{fig-2d} also suggest that the consistency improves and the customization declines with the margin increase, reflected by the CLIP score with the overall text gradually rising while the CLIP score with the view text decreasing.

\subsection{Text-to-3D Generation}
\label{sec-exp-3d}
In this section, we present text-to-3D generation results with DreamView-3D and compare them with other methods, where all methods are implemented based on the open-source threestudio library~\cite{threestudio}.

\begin{figure}[t]
  \centering
  \includegraphics[width=1\linewidth]{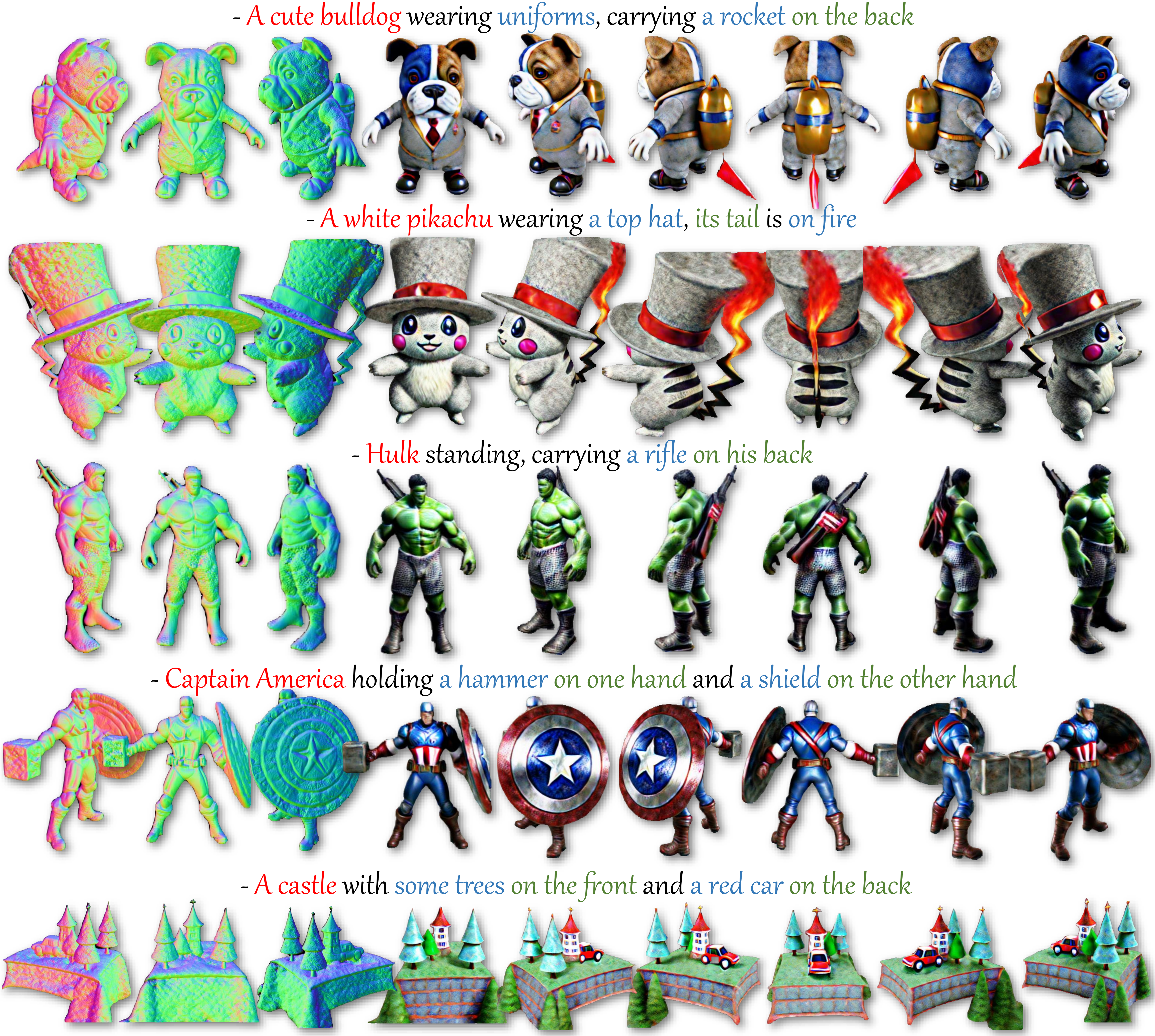}
  \caption{Text-to-3D generation of DreamView-3D. The first three columns present the rendered normal maps, and the rest are RGB images. We highlight the \textcolor{red}{subject}, \textcolor{bb}{object}, and the \textcolor{gg}{position} of the object in \textcolor{red}{red}, \textcolor{bb}{blue}, and \textcolor{gg}{green}. Only the overall texts are shown in the figure, and detailed view-specific texts are in \textit{supplement material}.}
  \label{fig-3d}
\end{figure}

\begin{figure}[!ht]
  \centering
  \includegraphics[width=1\linewidth]{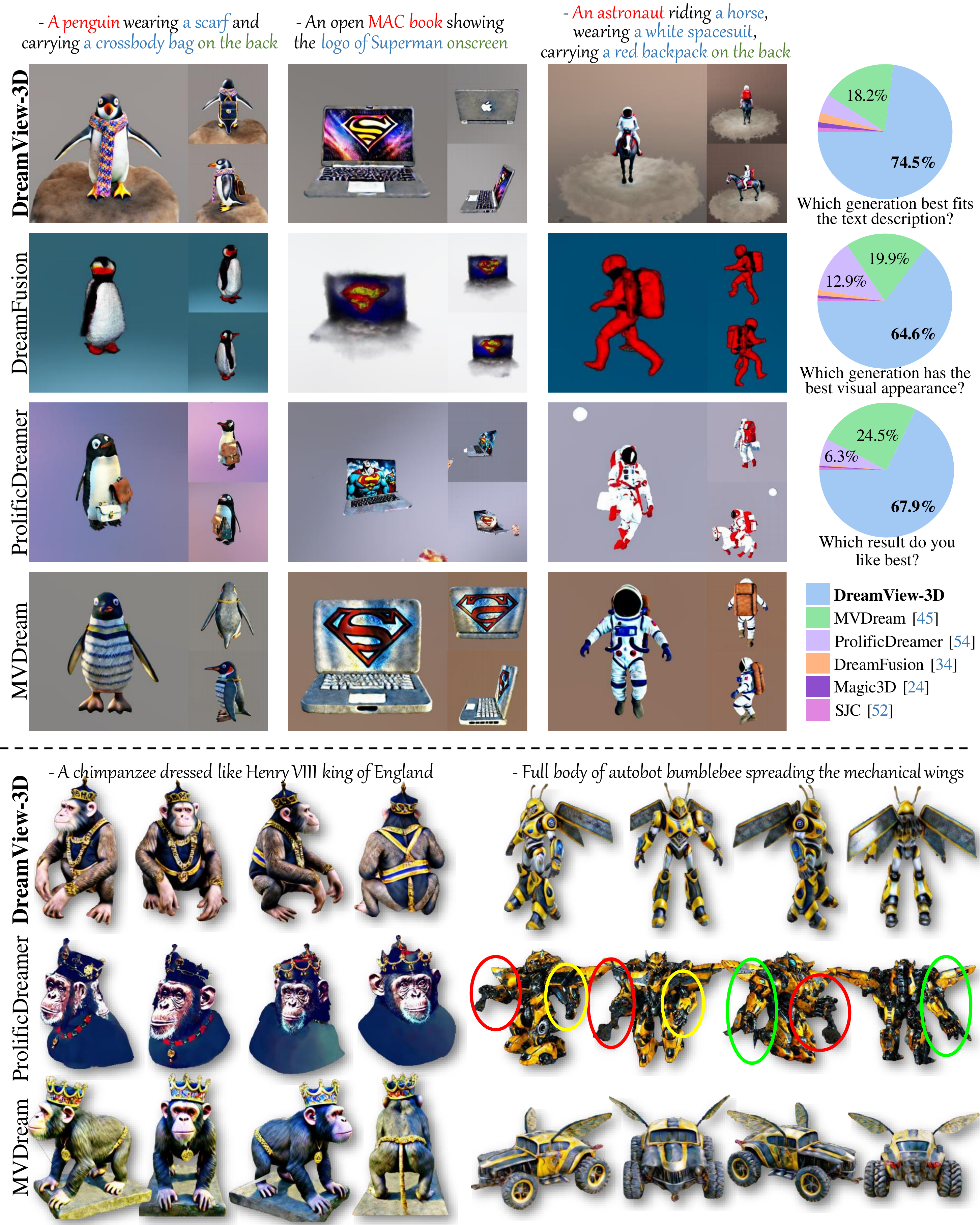}
  \vspace{-1.8em}
  \caption{\textbf{Top-left}: Comparisons with other methods on text-to-3D generation with customization requirements. Three views are shown, including the object's front, back, and side views. Other methods either overlook certain contents in the text prompts or fail to generate specific content in the expected position. Besides, some methods suffer from the 3D inconsistency problem. More comparisons are in the \textit{supplement material}. \textbf{Top-right}: The results of the user study. \textbf{Bottom:} Comparisons with other methods on text-to-3D generation with general text prompts (without viewpoint customization). The circles with three different colors show that the bumblebee has three arms.}
  \label{fig-comparison}
\end{figure}

\vspace{0.7em}
\noindent\textbf{Qualitative results.}
In~\Cref{fig-3d}, we show several qualitative results of our proposed DreamView-3D, where all contents specified in the text are accurately presented at their expected location, \eg, the rocket on the bulldog's back, the fire on the Pikachu's tail. Except for customizing from the front and back views, DreamView can also manipulate content from the left and right views, \ie, the hammer and shield on Captain's hand. Moreover, DreamView-3D can not only customize character-like objects but also works with scene objects, as shown in the last sample of~\Cref{fig-3d}, where a castle with trees and a red car is presented. In addition, our method not only enjoys an impressive customizable property but also maintains instance-level consistency, where none of the results show the multi-face and multi-foot problems. Overall, our DreamView-3D enables view customization while ensuring 3D consistency, thereby paving the way for generating more imaginative 3D assets.

Furthermore, we compare DreamView-3D with other methods on text-to-3D synthesis requiring customization in~\Cref{fig-comparison} (top-left), where we show the generated objects' front, back, and side views. According to the results, all the compared methods successfully generate the main concepts, such as the penguin, the MAC book, and the astronaut. However, in the case of generating the penguin, except for our DreamView-3D, other methods either ignore the objects (the crossbody bag and the scarf) or fail to place the bag in the expected location. Besides, other methods suffer from 3D inconsistency to some extent. For example, they always tend to generate the canonical view of the object, \ie, the face of the penguin and the screen of the MAC book. Despite MVDream being designed for 3D consistency, we still find that in the second example, the back of the MAC book shows the Superman logo instead of the Apple one. Moreover, in the last case, DreamFusion and MVDream fail to generate the horse, and ProlificDreamer only shows the horse in the side view. In comparison, ours can correctly generate all described contents and place each in the expected position. 

On the other hand, in the bottom side of~\Cref{fig-comparison}, we compare DreamView-3D with other methods on general prompts, \ie, no customization is required. The results demonstrate that our model can also work with general prompts and generate highly consistent 3D objects with high fidelity. Note that although ProlificDreamer~\cite{pdreamer} generates more complex and rich details, it severely suffers from the inconsistency problem, \eg, generating multiple faces for the chimpanzee and multiple arms for the bumblebee. On the last row, despite the specification of `full-body', MVDream~\cite{mvdream} still generates the car form of the bumblebee, which is not as user's expected. The above results show that our model can still work correctly even if only the overall text is provided.

\vspace{0.7em}
\noindent\textbf{User study.} 
We conducted a user study to collect user preferences and to provide an intuitive and comprehensive evaluation. We collected 30 text prompts and applied 6 different methods to generate 180 3D objects in total. Among these 30 text prompts, 15 of them have customization requirements, thus including both the overall text and the view-specific text, and the other 15 are general text prompts. The compared methods include DreamFusion~\cite{dreamfusion}, SJC~\cite{sjc}, Magic3D~\cite{magic3d}, ProlificDreamer~\cite{pdreamer}, MVDream~\cite{mvdream}, and ours. Each group of generation results is presented in the form of \{overall text, view-specific text (if used), results of 6 methods in random order\}. Each user is asked to answer three choice questions for each group of generations: (1) Which result best fits the text description? (2) Which result has the best visual appearance? (3) Which result do you like best? The evaluation criteria for these three questions are: (1) Are all the concepts in the text description presented correctly in the results? (2) Is the geometry complete or includes multi-face, multi-foot, or noise? Moreover, question (3) is subjective. 

In our user study, 35 participants with varying expertise and aesthetic views are involved, 25 are online volunteers, and the remaining 10 are researchers of related fields, \eg, 3D modeling, computer graphics, \etc. We received 1,050 feedbacks in total, and the results are shown in~\Cref{fig-comparison} (top-right). According to the results, 74.5\% of users choose our DreamView-3D to be more capable of generating 3D assets consistent with text descriptions, which is much higher than other methods. Regarding the quality of visual appearance, users choosing MVDream and ProlificDreamer increased, but it is still less than choosing ours. For the last question, 67.9\% users prefer ours over other methods, demonstrating the superior quality of our DreamView-3D.

\vspace{0.7em}
\noindent\textbf{Generation speed.} 
DreamView-3D takes roughly 55 minutes to generate a 3D asset on a single A100 GPU. Under the same experimental environment, DreamFusion~\cite{dreamfusion}, Magic3D~\cite{magic3d}, and SJC~\cite{sjc} take around 30 minutes, MVDream~\cite{mvdream} takes about 50 minutes, and ProlificDreamer~\cite{pdreamer} takes $\sim$180 minutes.

\section{Limitations and Conclusions}
\noindent\textbf{Limitations.}
The generated full-body characters' faces may be blurry and lose details, as shown by the third and fourth rows in~\Cref{fig-3d}, probably caused by using low-resolution training images. Training a higher-resolution model may address this problem but requires more training resources and time. (2) Besides, despite supposing customization, DreamView requires texts from different viewpoints to describe the same instance. Otherwise, the generation will fail, \eg, generating a dog from the front while a monkey from the back.

\vspace{0.7em}
\noindent\textbf{Conclusions.}
In this work, we introduce DreamView, a text-to-image model that can be lifted to 3D object generation and enables viewpoint customization while maintaining instance-level consistency by collaborating view-specific text and overall text via an adaptive guidance injection module. Extensive quantitative and qualitative results demonstrate the advancement of our method in text-to-3D generation, where our DreamView provides a highly versatile and personalized avenue for producing consistent and customizable 3D assets. 

\vspace{0.7em}
\noindent\textbf{Acknowledgement.}
This work was partially supported by the Guangdong NSF Project (No. 2023B1515040025, 2020B1515120085) and NSFC (U21A20471, U1911401), and the Guangdong Basic and Applied Basic Research Foundation (2023A1515012974). 
This work was also supported by Alibaba Group through the Alibaba Innovative Research Program. We also thank Kun-Yu Lin, Yi-Xing Peng, and Yu-Ming Tang for their helpful discussions.

\bibliographystyle{splncs04}
\bibliography{main}

\clearpage
\def\Ie{\emph{I.e}\onedot}
\def\cf{\emph{c.f}\onedot} \def\Cf{\emph{C.f}\onedot}
\def\etc{\emph{etc}\onedot} \def\vs{\emph{vs}\onedot}
\def\wrt{w.r.t\onedot} 
\def\aka{a.k.a\onedot} 
\def\dof{d.o.f\onedot}
\def\etal{\emph{et al}\onedot}
\def\ie{\textit{i.e.}}
\def\eg{\textit{e.g.}}
\renewcommand\thetable{S\arabic{table}}
\renewcommand\thefigure{S\arabic{figure}}
\renewcommand\thesection{S\arabic{section}}
\definecolor{gg}{RGB}{84, 130, 53}
\definecolor{pp}{RGB}{153, 51, 255}
\definecolor{bb}{RGB}{53, 121, 184}

\title{\textit{Supplementary Materials for}\\DreamView: Injecting View-specific Text Guidance into Text-to-3D Generation} 
\titlerunning{ }
\author{ }
\authorrunning{ }
\institute{ }
\maketitle

In this supplementary material, we provide the details of the dataset we rendered and our implementations (\Cref{ss.1}). Additionally, we present more qualitative text-to-3D generation results in~\Cref{ss.2}, including more comparisons, more results with general prompts, and combining DreamView-2D with another 2D-lifting technique. And we also discuss directly applying our injection module to other text-to-image models for 3D generation and some other ways to exploit the view-specific guidance. Besides, we discuss two potential ways to reduce the users' burden to write view-specific prompts in~\Cref{ss.3}. The dataset, model, and code will be released at \href{https://github.com/iSEE-Laboratory/DreamView}{here}.

\section{More Implementation Details}
\label{ss.1}
\noindent\textbf{Rendering Details.}
Our dataset is rendered from the Objaverse~\cite{objv} dataset, which contains more than 800k 3D assets. We follow the following pipeline to construct our dataset.
\begin{enumerate} 
    \item \textbf{Rendering}: For each 3D asset, we first normalize the object at the center covered by a bounding box between $[-0.5,0.5]$. Then, we random sample a camera elevation between $[0,30]$ and adjust the camera distance to ensure the entire 3D object is visible at any azimuth. Last, we uniformly render 32 images for the 3D object, starting from azimuth=0, and the camera position is saved at the same time. The illuminant comes from a three-point light source. The rendered size is set to be $512\times512$, and the background of the rendered image is a gray background. In total, we render $\sim$435k 3D assets, resulting in $\sim$14M rendered images.
    
    \item \textbf{Captioning}: We apply the BLIP-2~\cite{blip2} captioning model to caption the rendered images. For each image, we generate 5 captions. Then, we use CLIP~\cite{clip} to select one caption with the highest CLIP score with the rendered image, forming the view-specific text. 
    
    \item \textbf{Merging}: We use GPT-4~\cite{gpt4} to merge the 32 view-specific text, forming the overall text. The input of GPT-4 is a sentence templated as ``\textit{Given a set of descriptions about the same 3D object, distill these descriptions into one concise caption. The descriptions are as follows: `{text}'. Avoid describing the background. The caption should be:}'', where the `{text}' is the used 32 texts.
\end{enumerate}

\begin{figure}[t]
    \centering
    \includegraphics[width=1\linewidth]{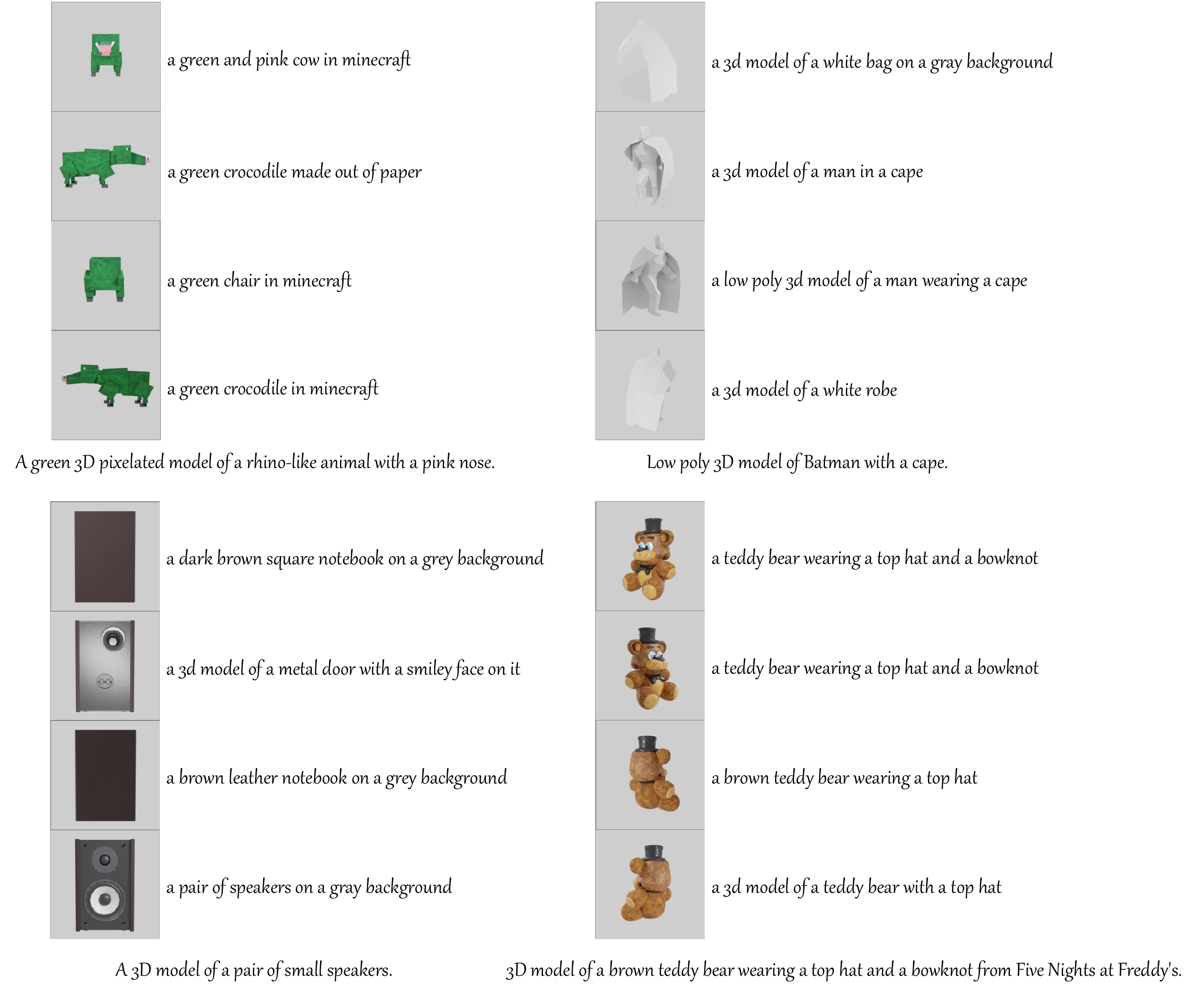}
    \caption{Samples from our dataset. We render 32 views for each 3D object in the Objaverse~\cite{objv} dataset and 4 orthogonal image views are shown in this figure. The text beside the image is its corresponding view-specific text and the text below the image is the overall text.}
    \label{fs.0}
\end{figure}

\noindent
We show some samples from our dataset in~\Cref{fs.0}. As we can see, the view-specific text does describe the content visible from its corresponding viewpoint. For example, only the cape (robe) can be seen in the sample's first and fourth image views in the top-right corner, so their texts only describe the cape (robe, bag). Moreover, the overall text integrates text descriptions from other viewpoints, thus providing a more comprehensive description, \eg, the ``Batman'' and ``from Five Nights at Freddy's'' in the top-right and bottom-right samples.

\vspace{1em}
\noindent\textbf{DreamView-2D.}
\begin{enumerate} 
    \item \textbf{Training}: We use a combined dataset consisting of our rendered dataset and a subset of the 2D LAION dataset~\cite{laion5b}. In each training iteration, we select 3D data with a 70\% probability and 2D data with a 30\% probability. For each 3D object, we randomly select 4 orthogonal image views, and their captions are suffixed with ``3D asset'' if the raw captions do not contain the word `3D' to distinguish from 2D data. We use 16 V100 to train the model. The batch size is 256 (4 objects, each object has 4 image views, across 16 GPUs), and the gradient is accumulated from 8 batches, forming a total batch size of 2,048. We use the AdamW~\cite{adamw} optimizer, and the learning rate is set to 1e$^{-4}$. The image size is $256\times256$. We randomly drop the text condition with a 10\% probability. The margin is randomly sampled from -0.1 to 0.1 for each training iteration. We use SD-v2.1 as the model initialization, and the base network architecture is mostly the same as those used in~\cite{dreamfusion,mvdream,pdreamer,sjc}.
    
    \item \textbf{Inference}: We adopt the DDIM~\cite{ddim} sampler with 50 sampling steps and a classifier-free guidance (CFG) scale of 7.5 for the text-to-image generation. By default, we generate 4 views simultaneously in inference and adopt a margin of -0.025.
\end{enumerate}

\vspace{1em}
\noindent\textbf{DreamView-3D.}
We implement DreamView-3D by building upon threestudio~\cite{threestudio} and substituting Stable Diffusion~\cite{sdm1} in DreamFusion~\cite{dreamfusion} with our text-to-image model (DreamView-2D) for text-to-3D generation. To represent the 3D content, we employ the implicit-volume approach (Mip-Nerf~\cite{barron2021mip}) and optimize it for 10,000 steps using the AdamW optimizer~\cite{adamw} with a learning rate of 0.01. The optimization takes about 55 minutes on an A100 GPU. During the first 8,000 optimization steps, SDS's maximum and minimum time steps are linearly decreased from 0.98 to 0.5 and 0.02, respectively. The CFG scale is set to 50, and we use a rescaling factor of 0.5 for the CFG rescaling. Initially, the rendering resolution and batch size are set to $64\times64$ and 8 for the first 5,000 steps and are then changed to $256\times256$ and 4. After 5,000 steps, we enable soft shading~\cite{magic3d}. In most cases, we divide the azimuth angle of the camera position into four intervals: $[10, 170]$ is the front side, $(170, 190)$ is the right side, $[190, 350]$ is the back side, and the remaining part is the left side. We use ``\textit{ugly, bad anatomy, blurry, pixelated obscure, unnatural colors, poor lighting, dull, and unclear, cropped, lowres, low quality, artifacts, duplicate, morbid, mutilated, poorly drawn face, deformed, dehydrated, bad proportions}'' as the negative text prompt. The margin of the DreamView-2D is set to be -0.025.

\section{More Evaluations}
\label{ss.2}
\subsection{Comparisons with Others on Text-to-3D Generation}
\begin{figure}[ht]
    \centering
    \includegraphics[width=1\linewidth]{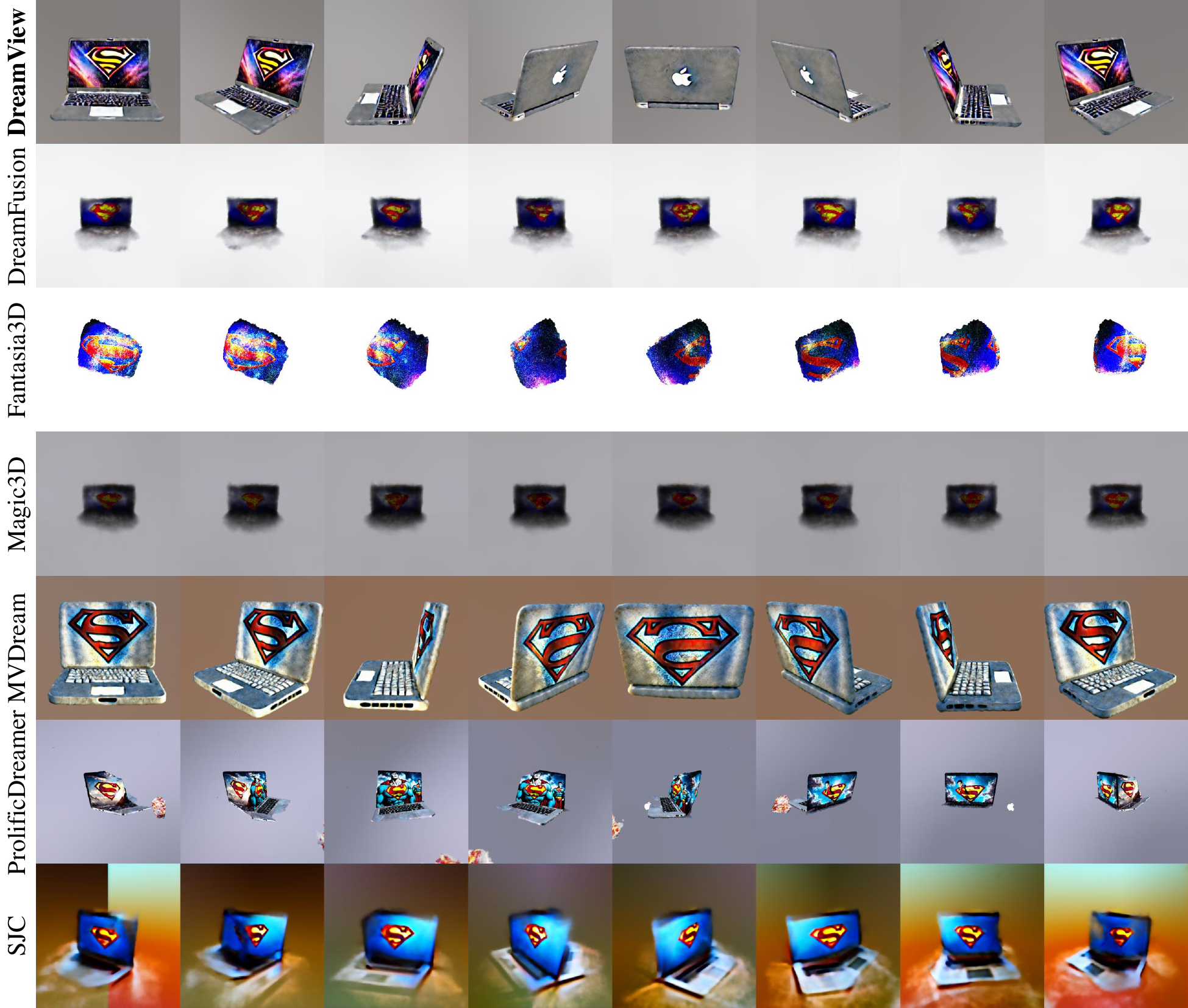}
    \caption{3D generation results of ``An open MAC book showing the logo of Superman on the screen''. View-specific text: -Front: ``An open MAC book showing the logo of Superman on the screen''. -Back: ``Back and side view of an open MAC book with an apple logo''. -Side: ``Side view of an open MAC book''. Note that the view-specific texts for the right and left are the same.}
    \label{fs.1}
\end{figure}

\begin{figure}
    \centering
    \includegraphics[width=1\linewidth]{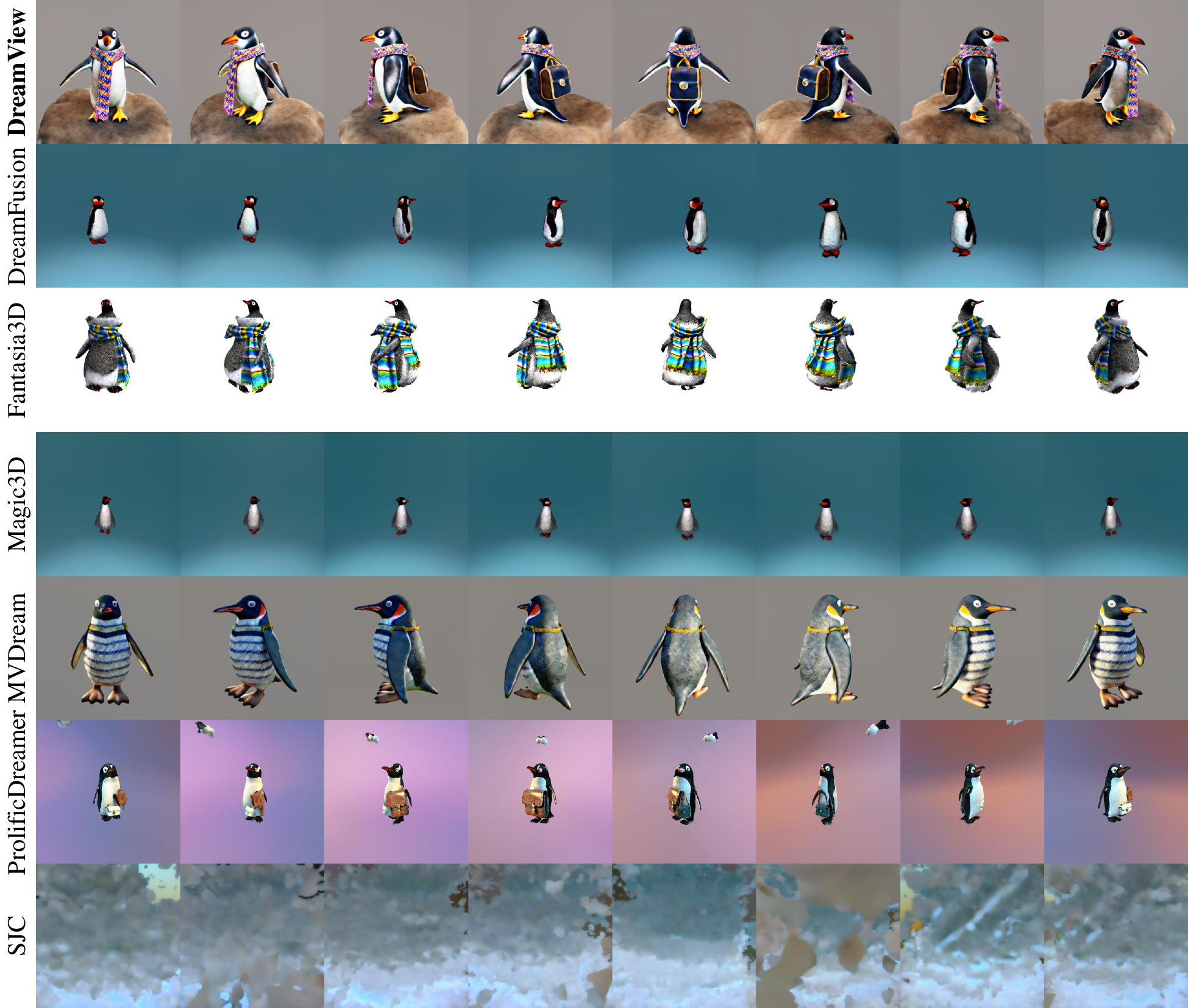}
    \caption{3D generation results of ``A penguin wearing a scarf, carrying a crossbody bag on the back, 8K, HD, 3d render, best quality''. View-specific text: -Front: ``A penguin wearing a scarf, 8K, HD, 3d render, best quality''. -Back: ``A penguin carrying a crossbody bag on the back, 8K, HD, 3d render, best quality''. -Side: ``A penguin wearing a scarf, carrying a crossbody bag on the back, 8K, HD, 3d render, best quality''. Note that the view-specific texts for the right and left are the same.}
    \label{fs.2}
\end{figure}

\begin{figure}
    \centering
    \includegraphics[width=1\linewidth]{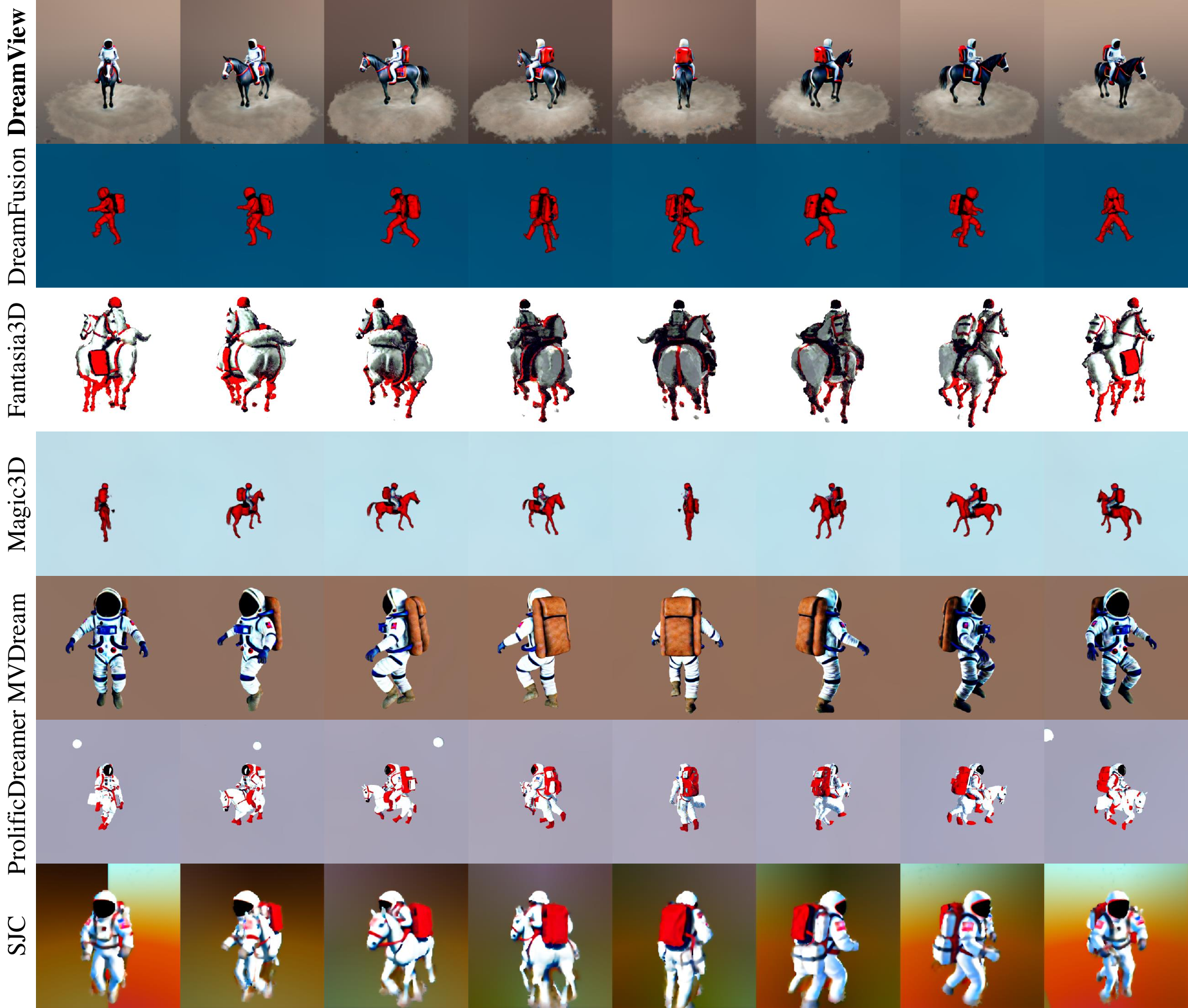}
    \caption{
    3D generation results of ``An astronaut riding a horse, wearing a white space suit, carrying a red backpack on the back''. View-specific text: -Front: ``A horse and an astronaut with a white space suit is riding the horse''. -Back: ``A red backpack is carried on the astronaut's back''. -Side: ``A red backpack, a horse, an astronaut with a white space suit''. Note that the view-specific texts for the right and left are the same.}
    \label{fs.3}
\end{figure}

In~\Cref{fs.1,fs.2,fs.3}, we compared the 3D generation results of 7 different methods under 3 text prompts containing customized content. We start from azimuth=0 degrees and show a total of 8 views at 45-degree intervals. These results demonstrate that other methods either suffer from the 3D inconsistent problem, \eg, multi-face and multi-foot problems, or cannot accurately generate specified content based on the text. In contrast, our method can include both 3D consistency and viewpoint customization capabilities.


\clearpage
\begin{figure}[t]
  \centering
  \includegraphics[width=1\linewidth]{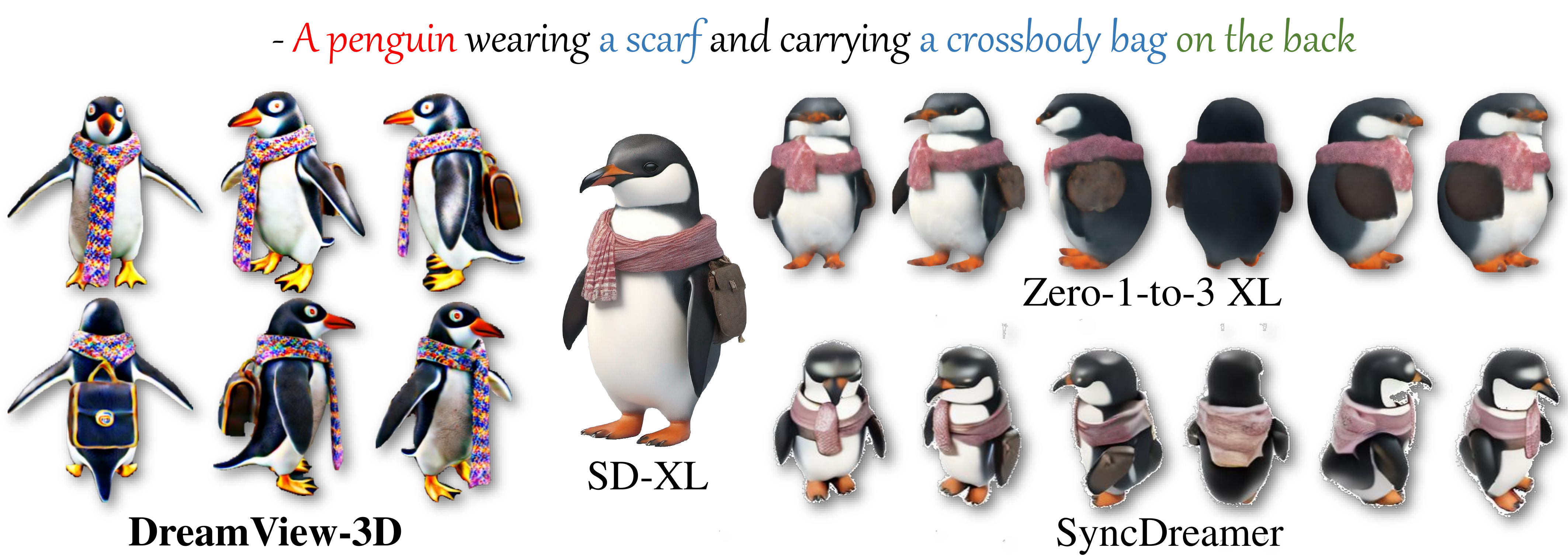}
  \caption{Comparisons between DreamView-3D and image-to-3D methods. The reference image for performing image-to-3D is generated by SD-XL with the overall text.}
  \label{fig-i23d}
\end{figure}

\subsection{Discussion with Image-to-3D Generation} 
In addition to text-to-3D generation, another stream of methods synthesizes 3D assets via lifting 2D images. We compare ours with them in~\Cref{fig-i23d}, where the reference image is generated by SD-XL using the overall text prompt. According to the results, although SD-XL can generate the concepts of the penguin, scarf, and crossbody bag, the crossbody bag is still overlooked when the generated image is lifted to a 3D object via Zero-1-to-3 XL~\cite{0123xl} and SyncDreamer~\cite{liu2023syncdreamer}. More importantly, these methods accept a single image for generating 3D assets, which inherently contradicts the diversity of appearance across viewpoints, making such singe-image-to-3D methods difficult to achieve customizable 3D generation. 

\clearpage
\begin{figure}[ht]
    \centering
    \includegraphics[width=1\linewidth]{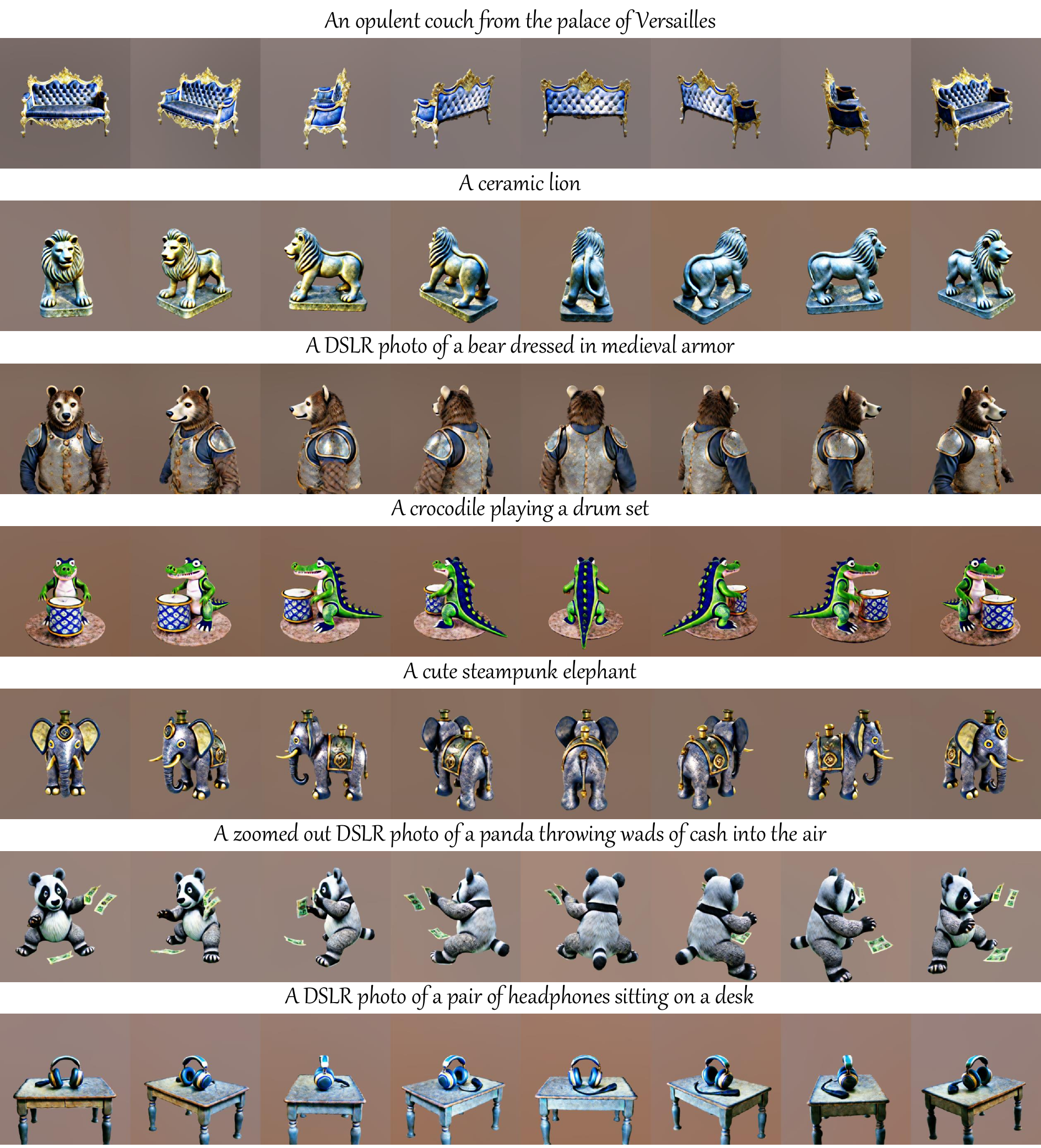}
    \caption{3D generation results of our DreamView with general prompts, \ie, only the overall text is used.}
    \label{fs.4}
\end{figure}

\subsection{More Results with General Prompts}
We also show more generation results with general prompts in~\Cref{fs.4}, where the used overall text prompts are from the prompt library~\footnote{\href{https://github.com/threestudio-project/threestudio/blob/main/load/prompt_library.json}{Prompt Library}}. The results demonstrate that our model can also work with general prompts and generate highly consistent 3D objects with high fidelity.

\clearpage
\subsection{Combing DreamView-2D with SJC Guidance}
\begin{figure}[t]
    \centering
    \includegraphics[width=1\linewidth]{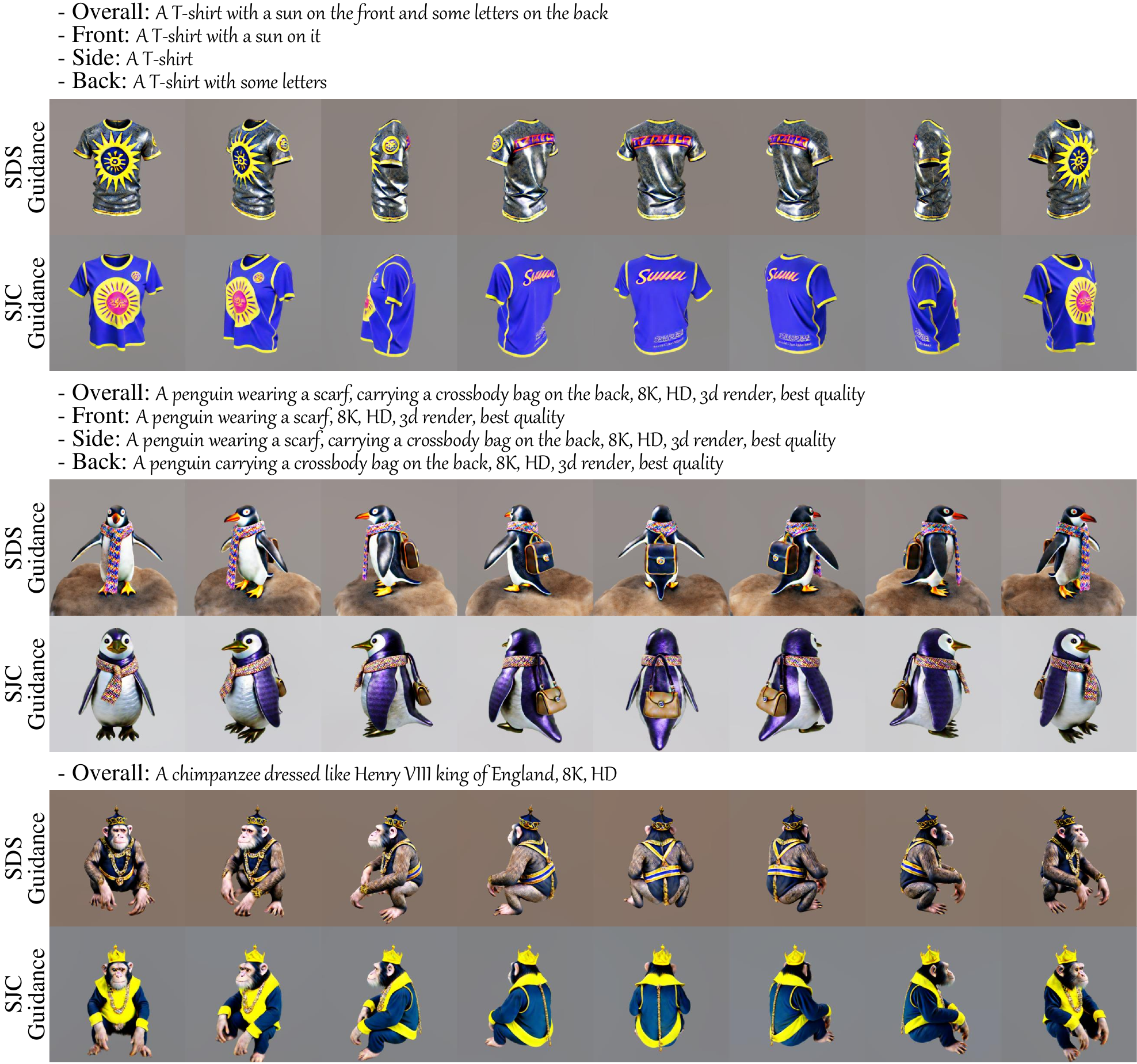}
    \caption{3D generation results via SDS guidance~\cite{dreamfusion} and SJC guidance~\cite{sjc}. The first two cases adopt view-specific texts, and the last case merely uses the overall text.}
    \label{fs.5}
\end{figure}
In the main manuscript, we mainly build our DreamView-3D upon the SDS guidance proposed in DreamFusion~\cite{dreamfusion}. Now, we adopt the SJC~\cite{sjc} guidance to lift our DreamView-2D to conduct text-to-3D generation, where the results are shown in~\Cref{fs.5}. The results demonstrate that DreamView-2D can work with other 2D-lifting techniques, such as SJC, not limited to SDS, and the properties of customization and consistency are still preserved. Besides, DreamView-2D is expected to be combined with VSD~\cite{pdreamer} to further improve the generation quality, which will be reserved for future work.

\clearpage
\subsection{Applying the Injection Module to Other Text-to-image Models}
\begin{figure}[t]
    \centering
    \includegraphics[width=1\linewidth]{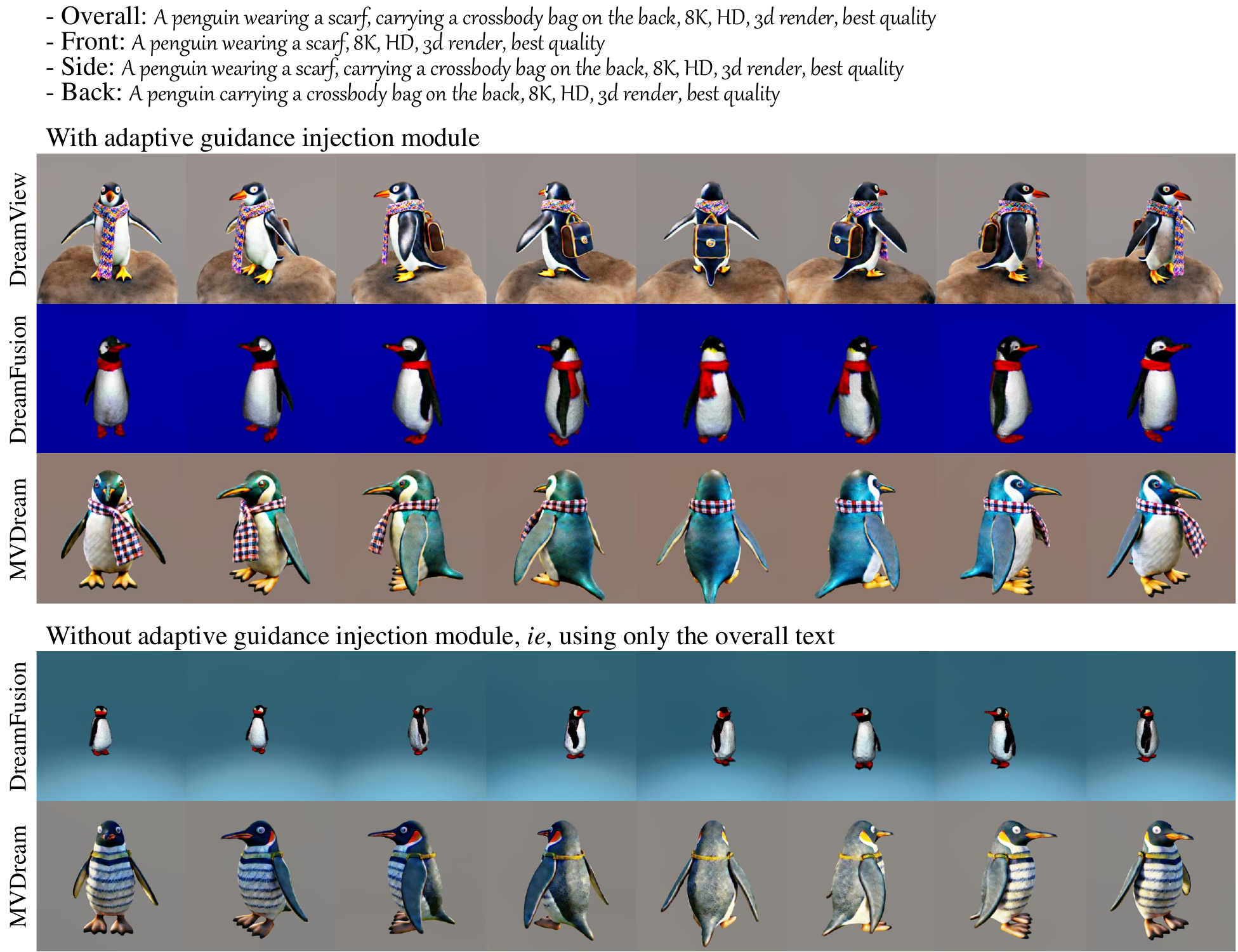}
    \caption{3D generation results of directly applying our adaptive injection module to DreamFusion and MVDream. The scarf is presented correctly for DreamFusion and MVDream with the injection module, but they still fail to generate the crossbody bag.}
    \label{fs.6}
\end{figure}
We directly apply our proposed adaptive guidance injection module to the text-to-image models used in DreamFusion~\cite{dreamfusion} and MVDream~\cite{mvdream} to verify if the injection module can introduce customization ability to them.

The results in~\Cref{fs.6} show that when the injection module is directly applied to DreamFusion, the scarf can be successfully presented while still failing to generate the crossbody bag. A similar phenomenon also occurs when applying the injection module to MVDream. 
Overall, our injection module can be directly integrated with these models and enables them to achieve some viewpoint customization without view customization training, verifying the universality of the injection method. 
However, their customization capability is not as ideal as our DreamView.

\clearpage
\subsection{Other Ways to Use View-specific Text Guidance}
\begin{figure}[t]
    \centering
    \includegraphics[width=1\linewidth]{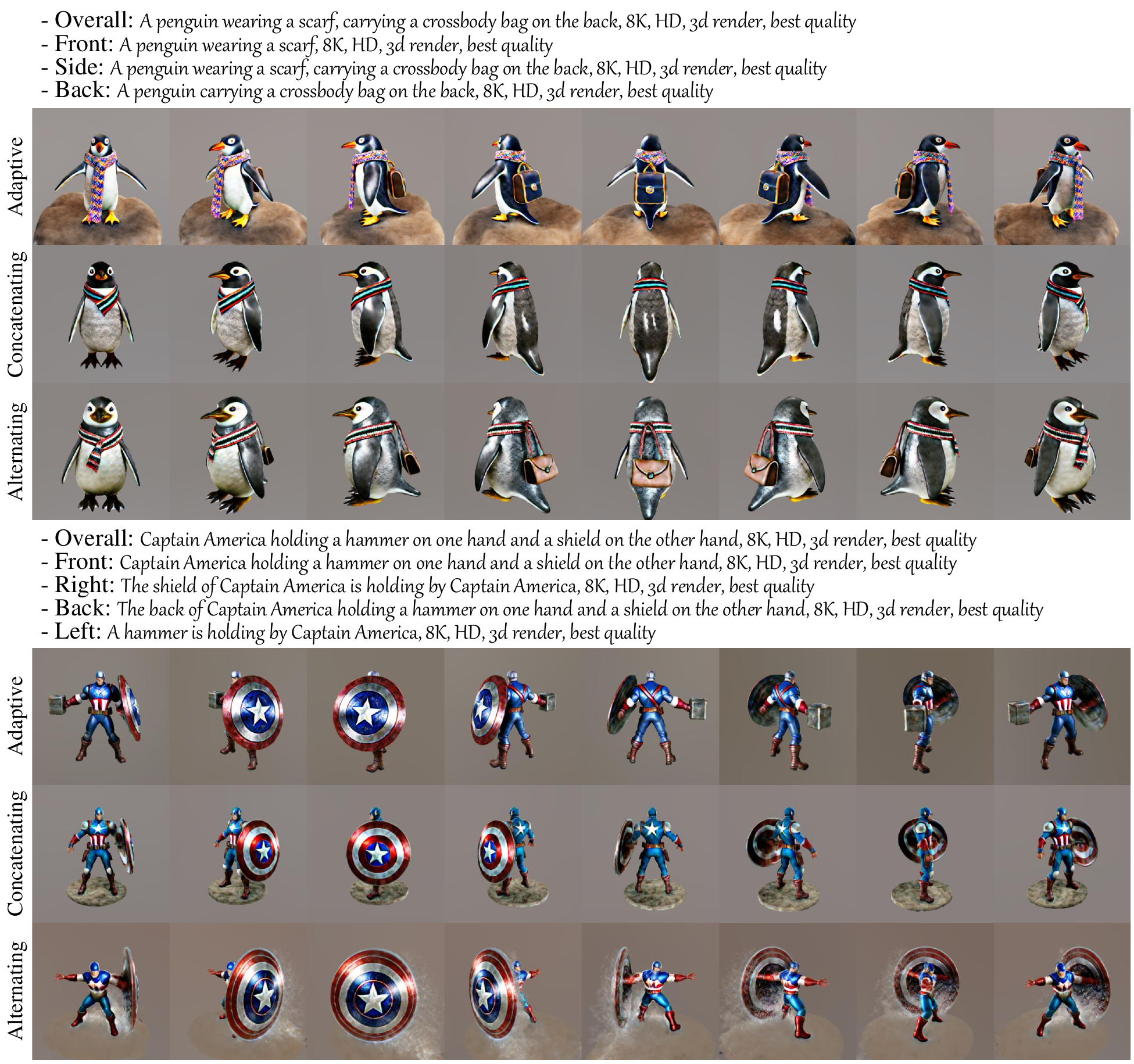}
    \caption{Comparisons between adaptive injection and another two injection methods.}
    \label{fs.7}
\end{figure}
We propose an adaptive way to use the view-specific and overall text guidance in DreamView. Except for adaptively injecting view-specific text guidance, we explore two other ways to inject them, termed \textbf{concatenating} and \textbf{alternating} injection. The former concatenates the text embeddings of the overall text and view-specific text to form the condition for all diffusion U-Net blocks. The latter alternately injects the overall text and view-specific text into the U-Net block. Specifically, it injects the view-specific text embedding to a U-Net block whose Layer ID can not be divided by 2. Otherwise, the overall text embedding is injected. In brief, these two injection methods inject the overall and view-specific text in a pre-defined static way. Compared with them, the adaptive injection module used in DreamView determines which guidance should be used in the current U-Net block according to the similarities between the text-image embedding, which is a dynamic process.

The comparisons are shown in~\Cref{fs.7}. 
According to the results, concatenating injection fails to fully customize the 3D content, \eg, the crossbody bag and the hammer. Besides, the alternating injection performs well in generating the penguin. However, it fails to generate the hammer in the second case, where we also observe slight 3D inconsistency, \ie, the multi-face problem and some noise. 

In conclusion, our adaptive injection module, which utilizes two kinds of guidance in a dynamic way, can achieve a better balance between consistency and customization than the concatenating and alternating injection methods that inject the overall and view-specific text in a static way.

\section{Reducing the Burden of Writing Prompts}
\label{ss.3}
All the text prompts used in DreamView, including the overall text and four view-specific prompts, are hand-written by default. Although such a way can ensure the accuracy of the description as much as possible, it undoubtedly brings additional burdens to users. Here, we discuss two potential ways to help users lower the burden of providing view-specific text prompts: (1) Appending the direction description text to the overall text to generate view-specific texts; (2) Asking Large Language Models (LLMs) for generating view-specific texts. We take the ``penguin'' case as an example and discuss them as follows.
\begin{figure}[ht]
    \centering
    \includegraphics[width=1\linewidth]{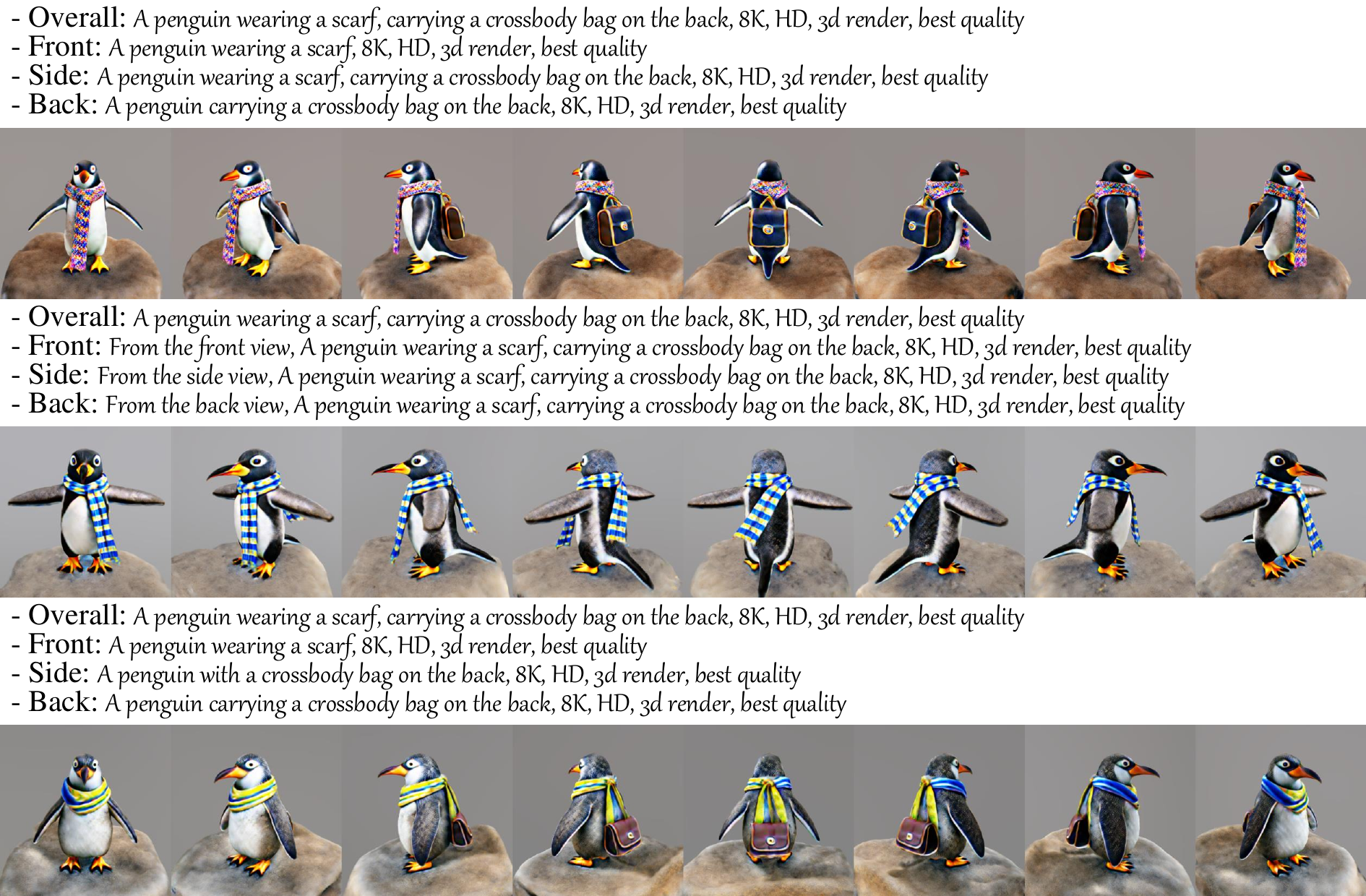}
    \caption{First row: 3D generation results of hand-written text prompts. Second row: 3D generation results of direction description text prompts. Last row: 3D generation results of text prompts given by GPT-4~\cite{gpt4}.}
    \label{fs.8}
\end{figure}

\vspace{1em}
\noindent\textbf{Way 1:}
We append the direction description text, \eg, `from the front view' and `from the back view', \etc, to the overall text to generate view-specific text, which is inspired by the view-dependent conditioning in DreamFusion~\cite{dreamfusion} that initially designed for alleviating the 3D inconsistent problem. The results are shown in the second row of~\Cref{fs.8}, where the crossbody bag is not successfully presented. We conjecture this is because the model has rarely learned the concept of direction description during its training. A possible solution is to add direction descriptions to the texts in the training set and retrain the model, but this may incur some annotation costs, which will be further explored in the future.

\vspace{1em}
\noindent\textbf{Way 2:}
We ask GPT-4~\cite{gpt4} to generate view-specific texts in a one-shot way, where the used text prompt template is:

\textit{The input is an overall text description of a 3D object from the global level. Please output its descriptions from the front, right, back, and left view based on this overall description, and return the result in a JSON format. I will give you an example:}

\textit{Input: `An Iron man carrying the shield of Captain America on the back'.}

\textit{Output:
\{`Front': `An Iron man',
`Right': `An Iron man with a shield on the back',
`Back': `The shield of Captain America is on the Iron man's back',
`Left': `An Iron man with a shield on the back'\}}

\textit{Now, what is the output of `A Penguin wearing a scarf, carrying a crossbody bag on the back, 8K, HD, 3D render, best quality'? Note: do not delete the text describing quality, eg, 8K, HD, and so on.}

\begin{table}[t]
  \caption{Comparisons on hand-written and GPT-given prompts.}
  \label{tab-r1}
  \centering
  {
    \begin{tabular}{c|c|c|c}
    \toprule
        How to obtain & \multicolumn{2}{c|}{CLIP Score ($\uparrow$)} & Inception \\ 
        \cline{2-3}
        View-specific text & Overall Text & View Text & Score ($\uparrow$) \\ 
        \hline
        Hand-written & 31.1 & 32.1 & 14.5 \\
        GPT3.5-given & 31.4 & 31.6 & 14.8 \\
    \bottomrule
    \end{tabular}
  }
\end{table}

The output text prompts and the 3D generation results are shown in the last row of~\Cref{fs.8}. The text prompts given by GPT-4 can successfully generate the penguin with a scarf and a crossbody bag, suggesting that using LLMs to generate view-specific texts may become a potential way to reduce user burden. Besides, Table~\ref{tab-r1} is the quantitative result of text-to-image generation on the validation set between hand-written and GPT3.5-given prompts, showing that the GPT-given prompts are quantitatively on par with hand-written prompts. 

\vspace{1em}
\noindent\textbf{A quick verification of prompts.} 
We here provide a quick verification of whether the used prompts can achieve satisfactory customization. In practice, users can refine their prompts according to the text-to-image generation results, as the quality of 3D generation is highly correlated with image generation. Since the image generation is much faster than the 3D generation (3 seconds \textit{vs.} 55 minutes), the prompt refinement process will not take long.


\end{document}